\documentclass{article} %
\usepackage{colm2024_conference}
\usepackage{booktabs}
\usepackage{siunitx}
\newcommand{\starnum}[1]{\num{#1}\rlap{\textsuperscript{*}}}
\newcommand{\ulnum}[1]{\underline{\num{#1}}}
\newcommand{\ulstarnum}[1]{\underline{\num{#1}}\rlap{\textsuperscript{*}}}

\usepackage{graphicx}
\usepackage{enumitem}
\usepackage{wrapfig}
\usepackage{algorithm}
\usepackage{algpseudocode}
\usepackage[T1]{fontenc}
\usepackage{lmodern} 
\usepackage{microtype}
\usepackage{amsmath}
\usepackage{colortbl}
\usepackage[utf8]{inputenc}
\definecolor{lightgray}{rgb}{0.9,0.9,0.9}
\usepackage{caption}
\usepackage{subcaption}
\usepackage{setspace}
\usepackage{url}
\usepackage{multirow}
\usepackage{colortbl}
\usepackage{tabularx}
\usepackage{blindtext}
\usepackage{pgfplots}
\pgfplotsset{compat=1.18} 
\usepackage{tikz}
\usetikzlibrary{er,positioning,bayesnet}
\usepackage{makecell}
\usepackage{tipa}
\usepackage{nicefrac}
\usepackage{tocloft}
\usepackage{listings}
\usepackage[raster,skins,breakable]{tcolorbox} %
\usepackage{xltabular}
\usepackage{adjustbox}
\usepackage{xurl}
\usepackage[normalem]{ulem}

\usepackage{amsmath,amsfonts,bm}

\def\eqref#1{equation~\ref{#1}}

\def\1{\bm{1}}

\DeclareMathAlphabet{\mathsfit}{\encodingdefault}{\sfdefault}{m}{sl}
\SetMathAlphabet{\mathsfit}{bold}{\encodingdefault}{\sfdefault}{bx}{n}

\newcommand{\snow}{\includegraphics[height=1em]{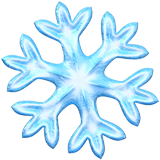}}
\newcommand{\fire}{\includegraphics[height=1em]{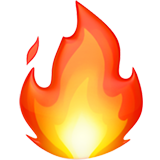}}

\title{Ovis2.5 Technical Report}

\author{
\bf Ovis Team, Alibaba Group
}

\begin{document}

\maketitle

\begin{abstract}

We present Ovis2.5, a successor to Ovis2 designed for native-resolution visual perception and strong multimodal reasoning. Ovis2.5 integrates a native-resolution vision transformer that processes images at their native, variable resolutions, avoiding the degradation from fixed-resolution tiling and preserving both fine detail and global layout—crucial for visually dense content like complex charts. To strengthen reasoning, we train the model to move beyond linear chain-of-thought and perform reflection—including self-checking and revision. This advanced capability is exposed as an optional ``thinking mode'' at inference time, allowing users to trade latency for enhanced accuracy on difficult inputs. The model is trained via a comprehensive five-phase curriculum that progressively builds its skills. The process begins with foundational visual and multimodal pretraining, advances through large-scale instruction tuning, and culminates in alignment and reasoning enhancement using DPO and GRPO. To scale these upgrades efficiently, we employ multimodal data packing and hybrid parallelism, yielding a significant end-to-end speedup. We release two open-source models: Ovis2.5-9B and Ovis2.5-2B. The latter continues the ``small model, big performance'' philosophy of Ovis2, making it ideal for resource-constrained, on-device scenarios. On the OpenCompass multimodal leaderboard, Ovis2.5-9B averages 78.3, marking a substantial improvement over its predecessor, Ovis2-8B, and achieving state-of-the-art results among open-source MLLMs in the sub-40B parameter range; Ovis2.5-2B scores 73.9, establishing SOTA for its size. Beyond aggregate scores, Ovis2.5 achieves leading results on STEM benchmarks, exhibits strong capabilities on grounding and video tasks, and achieves open-source SOTA at its scale for complex chart analysis.

\begin{figure}[H]
  \centering
  \includegraphics[width=\linewidth]{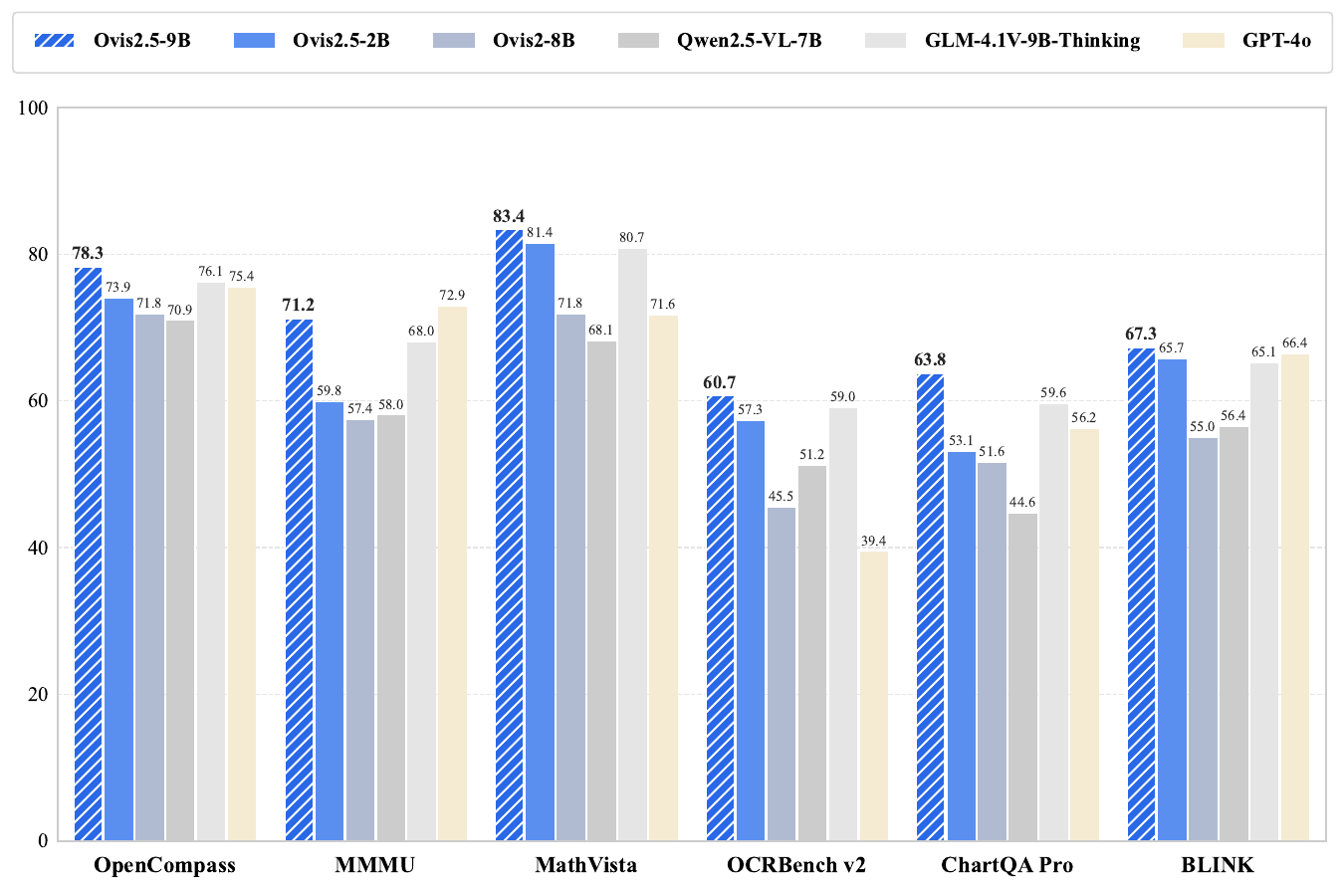}
  \caption{Benchmark performance of Ovis2.5 and its counterparts.}
  \label{fig:ovis25_bench}
\end{figure}

\end{abstract}

\vfill

\vfill

\section{Introduction}
\label{sec:intro}
The field of multimodal large language models (MLLMs) has seen rapid advancements, with a surge of powerful open-source models demonstrating remarkable capabilities \citep{liu2024llavanext, Qwen2.5-VL, zhu2025internvl3exploringadvancedtraining,glmvteam2025glm41vthinkingversatilemultimodalreasoning,kwaikeyeteam2025kwaikeyevltechnicalreport,coreteam2025mimovltechnicalreport}. Many of these models are built upon a conventional architecture, typically connecting a pre-trained vision transformer (ViT) and a large language model (LLM) via a simple projector like an MLP. In our previous work \citep{lu2024ovis}, we identified a misalignment between the continuous structure of the visual embeddings from the MLP projector and the discrete structure of the textual embeddings. We proposed the Ovis architecture to structurally align the textual and visual embeddings using a learnable visual embedding table, and empirically validated its superiority over conventional designs.

Building on this foundation, we have iteratively developed the Ovis series, releasing versions 1.5, 1.6, and 2.0. These updates progressively enhanced the model's capabilities to handle high-resolution images, multi-image and video inputs, OCR tasks, multilingual scenarios, and complex problems like mathematical reasoning. Each version demonstrated leading performance among open-source models of a similar scale upon its release. However, despite these significant improvements, we observed that even our latest model, Ovis2, still struggled with tasks requiring deep reasoning or detailed analysis of visually dense content, such as complex charts. These shortcomings stem from two core issues: (1) rigid vision front-ends—e.g., fixed-resolution encoders that necessitate image tiling and compromise global structure; and (2) training schemes that emphasize linear chain-of-thought (CoT) but lack reflective, self-corrective supervision, limiting deeper reasoning.

To address these issues, we present Ovis2.5, which introduces two key improvements:
First, to enhance its perceptual abilities, we replace the fixed-resolution ViT with the native-resolution ViT (NaViT,~\citealp{dehghani2023patchnpacknavit}). NaViT processes images at their native, variable resolutions, avoiding lossy tiling and preserving both fine-grained details and global layout that are critical for charts, diagrams, and other visually dense content. Second, to strengthen its reasoning abilities, we augment training with deep-reasoning data that supervises not only linear CoT but also reflective processes such as self-checking and revision. This teaches the model to produce intermediate steps to evaluate its own reasoning, and refine conclusions when necessary, enabling deeper and more robust reasoning. The deep-reasoning capability is available as an optional ``thinking mode'' at inference time. This allows users to keep the mode off for efficiency on easy inputs, or enable it for complex problems to trade latency for higher accuracy.

In alignment with these upgrades, we also refined our training strategy into a comprehensive five-phase curriculum—(i) visual pretraining, (ii) multimodal pretraining, (iii) instruction tuning, (iv) DPO~\citep{rafailov2023direct}, and (v) reinforcement learning (GRPO,~\citealp{shao2024deepseekmath})—which progressively builds the model from foundational perception to advanced reasoning. To scale this ambitious training, we developed a high-efficiency infrastructure that leverages multimodal data packing and hybrid parallelism, delivering a $3$–$4\times$ end-to-end speedup.

Comprehensive evaluations demonstrate the exceptional capabilities of Ovis2.5. On the OpenCompass multimodal leaderboard~\citep{duan2025vlmevalkitopensourcetoolkitevaluating}, Ovis2.5-9B achieves an average score of $78.3$, indicating strong overall multimodal performance. This result not only represents a substantial improvement over its predecessor, Ovis2-8B, but also establishes a new state-of-the-art among open-source MLLMs in the sub-40B parameter range. In line with Ovis2’s ``small model, big performance'' philosophy—particularly for on-device applications—we also release Ovis2.5-2B, which scores $73.9$ on OpenCompass, a state-of-the-art result among open-source MLLMs of comparable size. Beyond aggregate performance, Ovis2.5 demonstrates broad capability: it delivers leading results among open-source MLLMs of similar scale across STEM benchmarks and shows strong performance in grounding and video tasks. Notably, on complex chart-analysis benchmarks, Ovis2.5 sets a new state of the art for open-source models at its scale.

The key advances of Ovis2.5 are summarized as follows:
\begin{itemize}
\item \textbf{Native-Resolution Perception:} Integrates a native-resolution vision encoder to eliminate destructive tiling, preserving the crucial details and global context needed for superior analysis of complex visuals like charts and diagrams.
\item \textbf{Deep-Reasoning Capability:} Introduces an optional ``thinking mode'', enabled by training on data that demonstrates reasoning with reflection, which allows the model to move beyond linear CoT for deeper and more robust solutions.
\item \textbf{State-of-the-Art Performance:} Establishes a new SOTA on the OpenCompass leaderboard for open-source MLLMs in the $<40$B class and demonstrates leading performance across STEM, chart analysis, grounding, and video benchmarks.
\item \textbf{High-Efficiency Training:} Features an optimized training infrastructure powered by data packing and hybrid parallelism that delivers a $3$–$4\times$ end-to-end speedup.
\end{itemize}
\section{Architecture}

\begin{figure}[h]
  \centering
  \includegraphics[width=\linewidth]{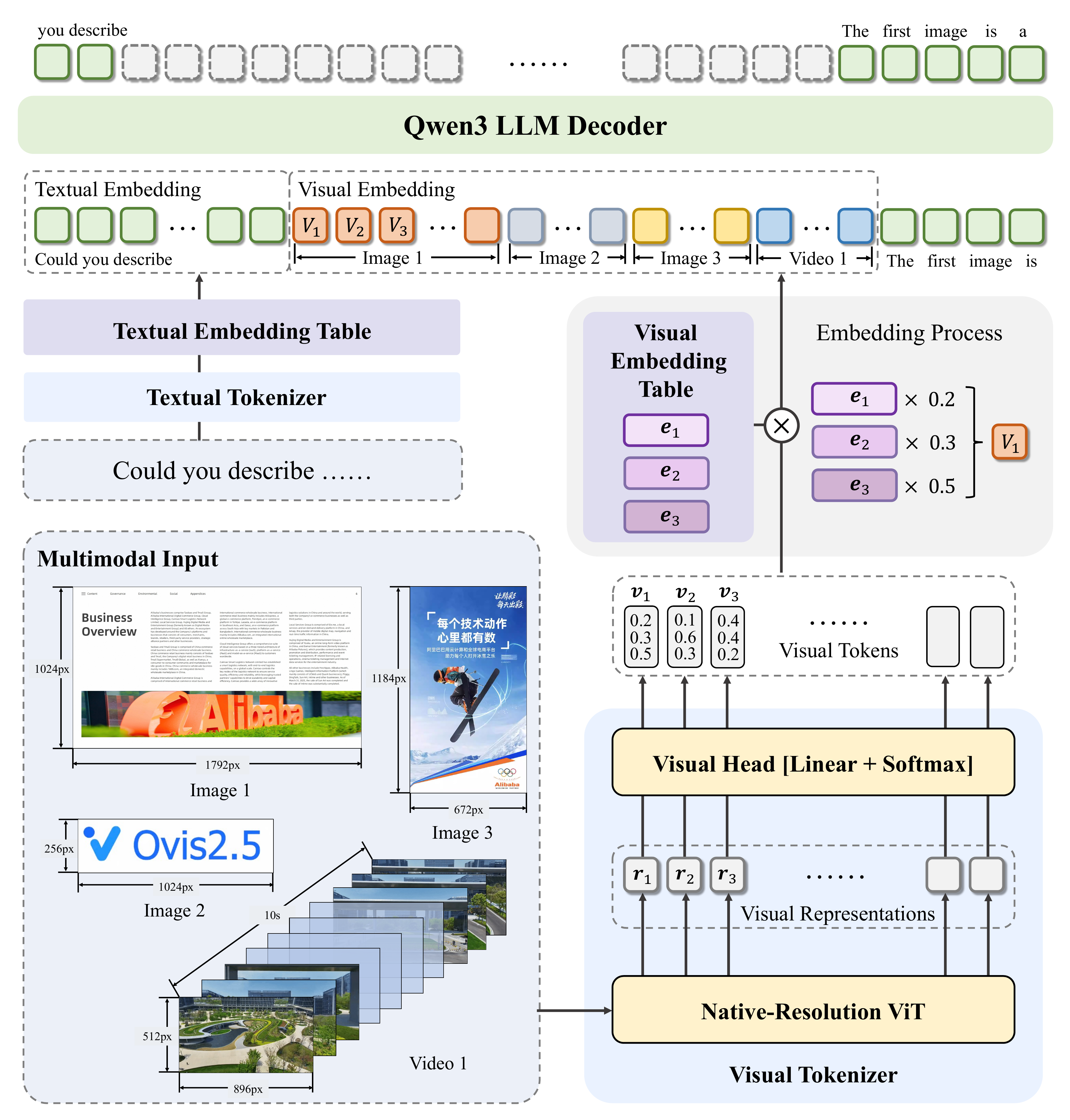}
  \caption{The overall architecture of Ovis2.5.}
  \label{fig:framework}
\end{figure}

The overall architecture of Ovis2.5 is illustrated in Figure~\ref{fig:framework}. Ovis2.5 retains the foundational design of the Ovis architecture~\citep{lu2024ovis}, comprising three core modules:

\begin{itemize}
    \item \textbf{Visual Tokenizer (VT):} A transformer-based component that extracts features from image patches. A visual head then projects each patch's features onto a discrete vocabulary of ``visual words'', producing a probability distribution over this vocabulary (i.e., a probabilistic visual token).

    \item \textbf{Visual Embedding Table (VET):} Analogous to a textual embedding table in LLMs, the VET stores a dedicated embedding for each visual word, a design which alleviates structural mismatches between modalities. The final visual embedding is computed by summing the table's embeddings, weighted by the probabilities from the VT. In other words, the embedding for a visual token is the expected value of the visual word embeddings, under the probability distribution produced by the VT.

    \item \textbf{LLM:} A pretrained open-source large language model that performs cross-modal understanding based on visual and textual embeddings to generate textual output.
\end{itemize}

To further advance Ovis2.5's multimodal capabilities, we have incorporated the following architectural enhancements:

\begin{itemize}
    \item \textbf{Native Resolution Processing:} Previous versions processed images by splitting them into fixed-size sub-images, which can disrupt global structure and fine-grained details. To overcome this, Ovis2.5 replaces the standard fixed-resolution ViT with a native-resolution ViT (NaViT), enabling the direct processing of images at their native resolutions. This architecture integrates rotary position embeddings (RoPE,~\citealp{su2024roformer}) into every ViT block to reinforce spatial awareness, proving especially effective for high-resolution images like complex charts. Our NaViT is initialized from the weights of siglip2-so400m-patch16-512\footnote{https://huggingface.co/google/siglip2-so400m-patch16-512}~\citep{tschannen2025siglip2multilingualvisionlanguage}.

    \item \textbf{Upgraded LLM:} We have replaced the Qwen2.5~\citep{an2024qwen2.5} backbone with Qwen3~\citep{qwen3}. This upgrade leverages the superior deep-reasoning capabilities of Qwen3 to significantly boost Ovis2.5's performance on complex tasks and its overall multimodal proficiency.
\end{itemize}
\section{Model Training}
The training pipeline for Ovis2.5 is structured into two primary stages: pre-training and post-training. The pre-training stage is subdivided into three distinct phases: Pre. Stage P1, Pre. Stage P2, and Pre. Stage P3. This is followed by a post-training stage, which contains two phases: Post. Stage P1 and Post. Stage P2. Each phase serves a distinct purpose, defined by its unique training configurations and data compositions. Table~\ref{tab:training} provides a high-level overview of this process, and the subsequent sections will detail each stage.

\begin{table}
  \caption{Overview of the Ovis2.5 training process.}
  \label{tab:training}
  \centering
  \small
  \begin{tabular}{c|cccc}
    \toprule
    \multirow{2}{*}{Phase} & \multicolumn{3}{c}{Modules} & \multirow{2}{*}{Data Composition} \\
    \cmidrule(lr){2-4}
                           & VT  & VET & LLM & \\
    \midrule
    Pre. Stage P1          & \snow/\fire & \fire & \snow & image caption \\
    Pre. Stage P2          & \fire       & \fire & \fire & OCR, image caption, grounding \\
    Pre. Stage P3          & \fire       & \fire & \fire & multimodal instruction dataset \\
    Post. Stage P1         & \fire       & \fire & \fire & multimodal preference dataset \\
    Post. Stage P2         & \snow       & \snow & \fire & multimodal RLVR dataset \\
    \bottomrule
  \end{tabular}
\end{table}

\subsection{Pre-training Stage}
\subsubsection{Data Composition}
The training of Ovis2.5 is underpinned by a large-scale, high-quality multimodal dataset, carefully assembled to support each phase of our training pipeline. This dataset is a curated collection, sourced from both public datasets and our in-house data. Public sources include a wide array of datasets such as COYO~\citep{kakaobrain2022coyo-700m}, Laion~\citep{schuhmann2022laion}, Wukong~\citep{gu2022wukong}, DataComp~\citep{gadre2023datacomp}, and SAM~\citep{kirillov2023segment}. To ensure data quality and diversity, we employ data processing pipelines tailored to each data type. Below, we outline several key components of the dataset.

\textbf{OCR Data.} Recognizing that robust OCR is a cornerstone of multimodal understanding, our corpus combines established public datasets with a diverse collection of in-house and web-sourced imagery, including documents, charts, posters, and screenshots. For this collected data, an ensemble of MLLMs is used to generate annotations and synthesize question–answer pairs. A subsequent curation step then filters samples based on image resolution, language, and scene diversity to ensure a high-quality, comprehensive OCR dataset.

\textbf{Grounding Data.} To equip the model with precise object localization capabilities, our grounding data is sourced from two primary sources. First, we leverage public datasets with existing bounding box annotations, such as RefCoCo~\citep{kazemzadeh2014referitgame}, to directly synthesize grounding instructions. Second, we develop an automated pipeline where state-of-the-art detection models identify and precisely localize entities within images, after which MLLMs generate corresponding question-answer pairs, effectively scaling our high-quality grounding data.

\textbf{Reasoning Data.} To cultivate sophisticated reasoning, we construct a diverse reasoning dataset. First, to establish the model's foundational reasoning ability, we assemble a large corpus of vanilla CoT data. We obtain this by collecting existing open-source CoT datasets and employing MLLMs to synthesize reasoning paths for QA pairs in our training set that originally contained only direct answers. Building on this foundation, and to encourage more complex cognition, we also use MLLMs to generate ``thinking-style'' data. This format, which explicitly incorporates \textless think\textgreater...\textless/think\textgreater ~tags, is designed to teach the model advanced cognitive processes such as reflection and self-correction. Finally, to guarantee the quality of all synthesized data, we implement a cross-verification labeling strategy, where multiple MLLMs assess and select the highest-quality samples for training.

\subsubsection{Training Strategy}
The pre-training process is divided into the following three phases:

\textbf{P1: VET Pre-training.} The primary goal of this phase is to train the Visual Embedding Table (VET). We utilize a dataset of image-caption pairs in a plain, non-dialogue format. To ensure stable learning, we initialize the ViT from siglip2-so400m-patch16-512 and freeze most parameters, limiting training exclusively to the final ViT layer, the visual head, and the VET. During this phase, input images are resized to a total pixel count between 448² and 896² while maintaining their original aspect ratio. Since the pre-trained ViT lacks Rotary Position Embeddings (RoPE), we employ dynamic position embedding interpolation to handle these varying resolutions, keeping RoPE itself disabled for stability.

\textbf{P2: Multimodal Pre-training.} This phase transitions to full-parameter training of all modules (VT, VET, and LLM) to establish core visual understanding and align the model with conversational formats. The training data is expanded to include conversational data from OCR, image captioning, and visual grounding tasks. Crucially, the supported resolution range is significantly expanded to a total pixel count between 448² and 1792² (approx. 200K to 3.2M pixels). To handle this increased spatial complexity, RoPE is activated in every ViT block, enhancing the model's spatial awareness.

\textbf{P3: Multimodal Instruction Tuning.} This phase continues full-parameter training, with a refined focus on enhancing the model's ability to follow diverse multimodal instructions. To this end, the training corpus is enriched with new input types—including text-only, multi-image, and video—and expanded to cover a wide spectrum of domains, such as general QA, multilingual dialogue, OCR, chart analysis, knowledge-based QA, STEM, and medical. To foster deeper reasoning beyond linear CoT, we incorporate ``thinking-style'' samples annotated with Qwen3-compatible \textless think\textgreater...\textless/think\textgreater ~tags, teaching the model to perform reflection and self-correction. The model continues to support the wide resolution range established in P2, allowing it to process high-resolution details.

\subsection{Post-training Stage}
\subsubsection{Data Composition}

\textbf{Direct Preference Optimization (DPO) Data.}
The DPO dataset covers multimodal preference data spanning text-only, single-image, multi-image, and video modalities. It consists of two main components:
(1) For reasoning-oriented tasks, we perform inference using both vanilla CoT and ``thinking-style'' formats and compute a verifiable score against ground-truth answers, thereby enhancing the model’s reasoning capabilities.
(2) For general-purpose tasks, we leverage an MLLM-based scoring mechanism to improve the model’s performance across diverse scenarios.

\textbf{Reinforcement Learning with Verifiable Rewards (RLVR) Data.}
Our RLVR dataset is collected from a wide range of domains. The primary focus is on open-source mathematical problems, with additional data including science QA and visual QA tasks involving logical reasoning. We also synthesize a substantial amount of data in which more information is embedded in images rather than in text. To improve the quality of multiple-choice questions, we implement a conversion process that transforms some of these questions into a \emph{fill-in-the-blank} format. This approach reduces the likelihood of correctly guessing answers without solid reasoning, a behavior that could harm training. The entire dataset undergoes both quality and offline difficulty filtering.

\subsubsection{Training Strategy}
The post-training process comprises the following two phases:

\textbf{P1: Multimodal DPO.}
In this phase, the entire Ovis2.5 model—including the vision modules and the LLM—undergoes full-parameter training. We use Direct Preference Optimization (DPO)~\citep{rafailov2023direct} as the primary preference objective, augmented with an auxiliary Negative Log-Likelihood (NLL) objective to stabilize optimization. For each query, a group of candidate responses is generated using the final model checkpoint from the pre-training stage. Within this group, we then form multiple preference pairs according to their labels.

\textbf{P2: Multimodal Reinforcement Learning.}
This phase further improves reasoning using Group Relative Policy Optimization (GRPO)~\citep{shao2024deepseekmath} on our RLVR dataset, building on the alignment established during the DPO phase. To focus optimization on high-level cognition while preserving general capabilities, we update only the LLM parameters and keep the vision modules frozen. Under this setup, GRPO optimizes the policy on verifiable, reasoning-centric tasks without degrading the model’s overall multimodal abilities.

\subsection{Infrastructure}
Efficiently training large multimodal models like Ovis2.5 presents significant infrastructure challenges. We focus on two primary issues: (1) computational load imbalance caused by the varying data sizes of images, videos, and text, and (2) memory limitations of our previous training framework that restrict model scale and context length. By developing targeted optimizations for both data processing and parallelism, we improve the end-to-end training speed of Ovis2.5 by 3 to 4 times.

\textbf{Load Balancing via Data Packing.}
A common inefficiency in training is padding, where samples of different lengths in a batch are padded to a uniform size. This leads to wasted computation and GPU idle time, especially with diverse multimodal data. To address this, we implement a data packing strategy. This method combines multiple shorter samples into a single, longer sequence during data preprocessing. As a result, it minimizes padding, reduces wasted computation, and creates a more balanced workload across GPUs, directly improving training throughput. Our experiments confirm that data packing significantly reduces training time while maintaining model accuracy.

\textbf{Hybrid Parallelism Framework.}
The increasing scale of our model architecture, particularly with the use of compute-intensive vision backbones, requires a more advanced parallelism strategy. We develop a hybrid parallelism framework for Ovis2.5 based on Megatron~\citep{shoeybi2019megatron}. This framework combines three standard techniques: Data Parallelism (DP), Tensor Parallelism (TP), and Context Parallelism (CP). This approach effectively reduces the memory footprint for large-scale model training, leading to substantial improvements in training throughput and efficiency.

\section{Experiments}
This section presents a comprehensive evaluation of Ovis2.5. We first compare its overall multimodal performance with leading MLLMs. Subsequently, we conduct a detailed assessment of its core capabilities, including multimodal reasoning, OCR and chart, visual grounding, and video understanding. Throughout this section, we use \textbf{bold} to denote the best and \underline{underline} for the second-best performance among open-source models of comparable scales. We use officially reported scores whenever available; otherwise, scores come from our own evaluation and are marked with an asterisk (*).

\begin{table}
  \caption{Performance of Ovis2.5-2B and comparison models on the OpenCompass suite. Abbreviations: MMB = MMBenchV11; MMS = MMStar; MMMU = MMMU-Val; HB = HallusionBench; OCR = OCRBench.}
  \label{tab:OpenCompass-2B}
  \centering
  \small
  \begin{tabular}{l|*{8}{S[table-format=2.1]}|S[table-format=2.1]}
    \toprule
    Model & {MMB} & {MMS} & {MMMU} & {MathVista} & {HB} & {AI2D} & {OCR} & {MMVet} & {Avg} \\
    \midrule
    Ovis2-2B           & 76.9 & 56.7 & 45.6 & 64.1 & 50.2 & 82.7 & 87.3 & 58.3 & 65.2 \\
    Qwen2.5-VL-3B      & 76.8 & 56.3 & \ulnum{51.2} & 61.2 & 46.6 & 81.4 & 82.8 & 60.0 & 64.5 \\
    InternVL3-2B       & 78.0 & 61.1 & 48.7 & 57.6 & 41.9 & 78.6 & 83.1 & 67.0 & 64.5 \\
    MiniCPM-V-4        & {\textbf{79.7}} & \ulnum{62.8} & \ulnum{51.2} & \ulnum{66.9} & \ulnum{50.8} & \ulnum{82.9} & {\textbf{89.4}} & {\textbf{68.0}} & \ulnum{69.0} \\
    \midrule
    Ovis2.5-2B         & \ulnum{79.2} & {\textbf{69.8}} & {\textbf{59.8}} & {\textbf{81.4}} & {\textbf{59.2}} & {\textbf{85.5}} & \ulnum{88.1} & \ulnum{67.9} & {\textbf{73.9}} \\
    \bottomrule
  \end{tabular}
\end{table}

\begin{table}
\caption{Performance of Ovis2.5-9B and comparison models on the OpenCompass suite. Abbreviations: MMB = MMBenchV11; MMS = MMStar; MMMU = MMMU-Val; HB = HallusionBench; OCR = OCRBench.}
  \label{tab:OpenCompass}
  \centering
  \small
  \begin{tabular}{l|*{8}{S[table-format=2.1]}|S[table-format=2.1]}
    \toprule
    Model & {MMB} & {MMS} & {MMMU} & {MathVista} & {HB} & {AI2D} & {OCR} & {MMVet} & {Avg} \\
    \midrule
    Gemini-2.5-Pro & 88.3 & 73.6 & 74.7 & 80.9 & 64.1 & 89.5 & 86.2 & 83.3 & 80.1 \\
    GPT-4o         & 86.0 & 70.2 & 72.9 & 71.6 & 57.0 & 86.3 & 82.2 & 76.9 & 75.4 \\
    \midrule
    Ovis2-8B             & 83.6 & 64.6 & 57.4 & 71.8 & 56.3 & 86.6 & {\textbf{89.1}} & 65.1 & 71.8 \\
    Qwen2.5-VL-7B        & 82.2 & 64.1 & 58.0 & 68.1 & 51.9 & 84.3 & \ulnum{88.8} & 69.7 & 70.9 \\
    InternVL3-8B         & 82.1 & 68.7 & 62.2 & 70.5 & 49.0 & 85.1 & 88.4 & \textbf{82.8} & 73.6 \\
    \midrule
    MiMo-VL-7B-RL-2508   & \starnum{83.9} & \starnum{72.7} & 70.6 & \starnum{79.7} & \ulstarnum{65.3} & \starnum{85.3} & 88.6 & \starnum{73.4} & \ulstarnum{77.4} \\
    Keye-VL-8B  & \starnum{79.4} & {\textbf{75.5}} & {\textbf{71.4}} & \ulnum{80.7} & {\textbf{67.0}} & 86.7 & 85.1 & \starnum{67.6} & \starnum{76.7} \\
    GLM-4.1V-9B-Thinking & {\textbf{85.3}} & \ulnum{72.9} & 68.0 & \ulnum{80.7} & \starnum{63.7} & {\textbf{87.9}} & 84.2 & \starnum{66.2} & \starnum{76.1} \\
    \midrule
    Ovis2.5-9B           & \ulnum{84.9} & 72.4 & \ulnum{71.2} & {\textbf{83.4}} & 65.1 & \ulnum{87.7} & 87.9 & \ulnum{74.0} & {\textbf{78.3}} \\
    \bottomrule
  \end{tabular}
\end{table}

\subsection{Overall Performance}
We evaluate the overall performance of Ovis2.5 using the OpenCompass suite~\citep{duan2025vlmevalkitopensourcetoolkitevaluating}, which provides a comprehensive score by averaging results across eight key benchmarks: MMBench~\citep{liu2024mmbench}, MMStar~\citep{chen2024mmstar}, MMMU~\citep{yue2024mmmu}, MathVista~\citep{lu2024mathvista}, HallusionBench~\citep{guan2024hallusionbench}, AI2D~\citep{kembhavi2016diagram}, OCRBench~\citep{liu2024ocrbench}, and MMVet~\citep{yu2023mmvet}. As presented in Tables~\ref{tab:OpenCompass-2B} and \ref{tab:OpenCompass}, with average scores of 78.3 and 73.9, respectively, Ovis2.5-9B and Ovis2.5-2B both achieve state-of-the-art performance among open-source models at their respective scales. Beyond these aggregate scores, the models demonstrate robust performance on general benchmarks such as MMBench and MMStar, and exhibit strong capabilities in several specialized domains.

Specifically, Ovis2.5-9B excels in complex reasoning tasks, attaining 71.2 on MMMU for college-level multi-discipline problems. It also scores 87.9 on OCRBench, underscoring its precise text recognition, and 65.1 on HallusionBench, indicating fewer hallucinations and higher reliability.

Collectively, these results demonstrate that Ovis2.5 not only surpasses leading open-source models like Keye-VL~\citep{kwaikeyeteam2025kwaikeyevltechnicalreport} and GLM-4.1V~\citep{glmvteam2025glm41vthinkingversatilemultimodalreasoning}, but also narrows the gap to frontier proprietary models such as Gemini-2.5~\citep{comanici2025gemini25pushingfrontier}. Complementing these cross-model comparisons, Ovis2.5 also delivers substantial gains over its predecessor, Ovis2, further validating the effectiveness of our architectural refinements and training strategy.

\subsection{Multimodal Reasoning}
We evaluate the multimodal reasoning capabilities of Ovis2.5 on a comprehensive suite of benchmarks—MMMU, MMMU-Pro~\citep{yue2024mmmupro}, MathVista, MathVerse~\citep{zhang2024mathverse}, MathVision~\citep{wang2024measuring}, LogicVista~\citep{xiao2024logicvistamultimodalllmlogical}, WeMath~\citep{qiao2024we}, and DynaMath~\citep{zou2024dynamic}—covering interdisciplinary academic reasoning, visual mathematics, logical reasoning, and dynamic problem solving.

As detailed in Table~\ref{tab:math_benchmarks_2b} and Table~\ref{tab:math_benchmarks}, Ovis2.5 excels across reasoning domains. Specifically, Ovis2.5-9B showcases leading capabilities in visual and structured mathematics among open-source models. It achieves the top rank on both MathVista and WeMath, demonstrating its exceptional strength in visual-compositional and concept-integration tasks. Its dominance is further solidified by consistently ranking among the top two on all other math-focused benchmarks, including MathVerse, MathVision, LogicVista, and DynaMath. Beyond its mathematical prowess, the model's proficiency in general academic reasoning is confirmed by strong scores of 71.2 on MMMU and 54.4 on the more challenging MMMU-Pro benchmark.
Notably, the compact Ovis2.5-2B delivers state-of-the-art results within its size category, achieving substantial gains over comparable open-source models across both general reasoning and math-oriented tasks.

\begin{table}
  \caption{Performance of Ovis2.5-2B and comparison models on multimodal reasoning benchmarks. Abbreviations: MPro = MMMU-Pro; MathVerse = MathVerse Vison Only; LV = LogicVista; WM = WeMath; DM = DynaMath.}
  \label{tab:math_benchmarks_2b}
  \centering
  \small
  \begin{tabular}{l|*{8}{S[table-format=2.1]}}
    \toprule
    Model & {MMMU} & {MPro} & {MathVista} & {MathVerse} & {MathVision} & {LV} & {WM} & {DM} \\
    \midrule
    Ovis2-2B           & 45.6 & 24.2   & 64.1 & 29.4 & 17.7 & 34.7 & 9.9  & 10.0 \\
    Qwen2.5-VL-3B      & \ulnum{51.2} & \ulnum{31.6} & 61.2 & \ulnum{31.2} & \ulnum{21.9} & \ulnum{40.3} & 22.9 & 13.2 \\
    InternVL3-2B       & 48.7 & {-}   & 57.6 & 24.5 & 20.2 & 34.7 & 22.9 & \ulnum{14.8} \\
    MiniCPM-V-4        & \ulnum{51.2} & {-}   & \ulnum{66.9} & {-}   & 20.7 & {-}   & \ulnum{32.7} & 14.2 \\
    \midrule
    Ovis2.5-2B         & {\textbf{59.8}} & {\textbf{39.3}} & {\textbf{81.4}} & {\textbf{64.8}} & {\textbf{37.4}} & {\textbf{53.0}} & {\textbf{50.8}} & {\textbf{26.5}} \\
    \bottomrule
  \end{tabular}
\end{table}

\begin{table}
  \caption{Performance of Ovis2.5-9B and comparison models on multimodal reasoning benchmarks. Abbreviations: MPro = MMMU-Pro; MathVerse = MathVerse Vison Only; LV = LogicVista; WM = WeMath; DM = DynaMath.}
  \label{tab:math_benchmarks}
  \centering
  \small
  \begin{tabular}{l|*{8}{S[table-format=2.1]}}
    \toprule
    Model & {MMMU} & {MPro} & {MathVista} & {MathVerse} & {MathVision} & {LV} & {WM} & {DM} \\
    \midrule
    Gemini-2.5-Pro & 74.7 & {-} & 80.9 & 76.9 & 69.1 & 73.8 & 78.0 & 56.3 \\
    GPT-4o         & 72.9 & {-} & 71.6 & 49.9 & 43.8 & 64.4 & 50.6 & 48.5 \\
    \midrule
    Ovis2-8B             & 57.4 & 34.9 & 71.8 & 42.3 & 25.9 & 39.4 & 27.2 & 20.4 \\
    Qwen2.5-VL-7B        & 58.0 & 38.3 & 68.1 & 41.1 & 25.4 & 47.9 & 36.2 & 21.8 \\
    InternVL3-8B         & 62.2 & \starnum{42.3} & 70.5 & 38.5 & 30.0 & 44.5 & 39.5 & 25.7 \\
    \midrule
    MiMo-VL-7B-RL-2508   & 70.6 & \starnum{45.7} & \starnum{79.7} & {\textbf{\starnum{71.6}}} & {\textbf{\starnum{58.5}}} & {\textbf{64.5}} & \ulstarnum{65.6} & {\textbf{\starnum{48.3}}} \\
    Keye-VL-8B  & {\textbf{71.4}} & \starnum{39.0} & \ulnum{80.7} & 59.8 & 46.0 & 54.8 & 60.7 & 37.3 \\
    GLM-4.1V-9B-Thinking & 68.0 & {\textbf{57.1}} & \ulnum{80.7} & \starnum{68.8} & \starnum{49.4} & \starnum{54.1} & 63.8 & \starnum{38.9} \\
    \midrule
    Ovis2.5-9B           & \ulnum{71.2} & \ulnum{54.4} & {\textbf{83.4}} & \ulnum{71.1} & \ulnum{53.9} & \ulnum{61.5} & {\textbf{66.7}} & \ulnum{44.1} \\
    \bottomrule
  \end{tabular}
\end{table}

\subsection{OCR \& Chart}
The challenge of interpreting visual data, such as documents and charts, demands a sophisticated cascade of skills from MLLMs, spanning from low-level OCR to high-level semantic reasoning. To rigorously evaluate these capabilities, we assess Ovis2.5 on a suite of challenging benchmarks.

On the large-scale, bilingual OCRBench v2~\citep{fu2024ocrbenchv2improvedbenchmark}, Ovis2.5 not only surpasses all leading open-source competitors but also outperforms the proprietary GPT-4o model. This state-of-the-art performance extends to complex chart analysis. We test our model on the newly introduced ChartQA Pro~\citep{masry2025chartqaprodiversechallengingbenchmark}, which features a diverse range of visualizations from conventional charts to complex infographics. Across these and other established benchmarks—including ChartQA~\citep{masry-etal-2022-chartqa}, DocVQA~\citep{mathew2021docvqa}, and TextVQA~\citep{singh2019towards}—our Ovis2.5-9B consistently demonstrates superior performance, attaining the top average score against its open-source peers, as detailed in Table \ref{tab:ocr_chart}.

\begin{table}
  \caption{Performance on OCR \& chart benchmarks. Abbreviations: OCRv2 = OCRBench v2; CQA = ChartQA; DocVQA = DocVQA-Val; TextVQA = TextVQA-Val.}
  \label{tab:ocr_chart}
  \centering
  \small
  \begin{tabular}{l|*{6}{S[table-format=2.1]}|S[table-format=2.1]}
    \toprule
    Model & {OCRv2-EN} & {OCRv2-CN} & {CQA Pro} & {CQA} & {DocVQA} & {TextVQA} & {Avg} \\
    \midrule
    GPT-4o                 & 46.5 & 32.2 & \starnum{56.2} & \starnum{85.8} & \starnum{90.9} & \starnum{89.6} & \starnum{66.9} \\
    \midrule
    Ovis2-8B               & 46.7 & 44.3 & 51.6 & 86.8 & 95.4 & {\textbf{92.5}} & 69.6 \\
    Qwen2.5-VL-7B          & 46.7 & 55.6 & \starnum{44.6} & \starnum{89.4} & \starnum{96.3} & \starnum{88.4} & \starnum{70.2} \\
    InternVL3-9B           & \starnum{46.2} & \starnum{44.1} & \starnum{50.2} & \starnum{87.4} & \starnum{93.7} & \starnum{90.2} & \starnum{68.6} \\
    \midrule
    MiMo-VL-7B-RL-2508     & \starnum{50.9} & \starnum{37.3} & \ulstarnum{62.6} & {\textbf{94.4}} & \ulstarnum{96.4} & \starnum{89.1} & \starnum{71.8} \\
    Keye-VL-8B-Thinking    & \starnum{45.0} & \starnum{36.2} & \starnum{46.9} & 86.3 & \starnum{88.6} & \starnum{87.7} & \starnum{65.1} \\
    GLM-4.1V-9B-Thinking   & \starnum{60.5} & \ulstarnum{57.4} & \starnum{59.6} & \starnum{87.9} & {\textbf{\starnum{97.1}}} & \starnum{91.0} & \ulstarnum{75.6} \\
    \midrule
    Ovis2.5-2B             & \ulnum{60.6} & 54.0 & 53.1 & 92.2 & 95.3 & 89.6 & 74.1 \\
    Ovis2.5-9B             & {\textbf{63.4}} & {\textbf{58.0}} & {\textbf{63.8}} & \ulnum{92.9} & 96.3 & \ulnum{91.2} & {\textbf{77.6}} \\
    \bottomrule
  \end{tabular}
\end{table}

\subsection{Grounding}
Visual grounding, the task of precisely localizing objects from natural language descriptions, is a cornerstone of advanced spatial reasoning in MLLMs. We benchmarked Ovis2.5 on the standard referring expression datasets: RefCOCO~\citep{kazemzadeh-etal-2014-referitgame}, RefCOCO+, and RefCOCOg~\citep{mao2016generation}. As detailed in Table \ref{tab:refcoco_benchmark}, Ovis2.5 achieves a state-of-the-art average score of 90.1 among compared open-source models.

This leading performance is consistent across individual benchmarks. The model delivers top-tier results across nearly all splits of RefCOCO and RefCOCO+, frequently ranking first or second. Crucially, Ovis2.5 demonstrates a pronounced advantage on the more demanding RefCOCOg dataset—a benchmark known for complex descriptions of non-salient objects. Here, it outperforms all competitors on both the validation and test sets. This exceptional performance, particularly on RefCOCOg, showcases the model's advanced understanding of visual-linguistic connections, proving its effectiveness for visual grounding tasks.

\begin{table}
  \caption{Performance on visual grounding benchmarks.}
  \label{tab:refcoco_benchmark}
  \centering
  \small
\begin{tabular}{l|*{8}{S[table-format=2.1]}S[table-format=2.1]}
  \toprule
  \multirow{2}{*}{Model}
  & \multicolumn{3}{c}{RefCOCO} & \multicolumn{3}{c}{RefCOCO+} & \multicolumn{2}{c}{RefCOCOg}
  & \multicolumn{1}{c}{\multirow{2}{*}{Avg}} \\
  \cmidrule(lr){2-4}\cmidrule(lr){5-7}\cmidrule(lr){8-9}
   & \multicolumn{1}{c}{val} & \multicolumn{1}{c}{test-A} & \multicolumn{1}{c}{test-B}
   & \multicolumn{1}{c}{val} & \multicolumn{1}{c}{test-A} & \multicolumn{1}{c}{test-B}
   & \multicolumn{1}{c}{val} & \multicolumn{1}{c}{test} &  \\
  \midrule
  InternVL3-8B         & 92.5 & {\bfseries 94.6} & \ulnum{88.0} & \ulnum{88.2} & {\bfseries 92.5} & \ulnum{81.8} & 89.6 & 90.0 & \ulnum{89.6} \\
  InternVL3-9B         & 91.8 & 93.2 & 86.6 & 86.4 & 91.0 & 79.9 & 88.0 & 88.5 & 88.2 \\
  Qwen2.5-VL-7B        & 90.0 & 92.5 & 85.4 & 84.2 & 89.1 & 76.9 & 87.2 & 87.2 & 86.6 \\
  \midrule
  MiMo-VL-7B-RL-2508   & {\bfseries \starnum{92.8}} & \starnum{93.8} & \starnum{86.7} & {\bfseries \starnum{89.2}} & \starnum{91.6} & \starnum{81.6} & \starnum{89.0} & \starnum{88.1} & \starnum{89.1} \\
  GLM-4.1V-9B-Thinking & \starnum{84.8} & \starnum{88.9} & \starnum{78.2} & \starnum{76.3} & \starnum{82.4} & \starnum{69.3} & \starnum{80.8} & \starnum{83.1} & \starnum{81.1} \\
  \midrule
  Ovis2.5-2B           & {\bfseries 92.8} & {\bfseries 94.6} & {\bfseries 89.7} & 86.9 & 91.0 & 81.5 & \ulnum{90.2} & {\bfseries 90.4} & \ulnum{89.6} \\
  Ovis2.5-9B           & \ulnum{92.7} & \ulnum{94.3} & {\bfseries 89.7} & 87.7 & \ulnum{92.1} & {\bfseries 83.6} & {\bfseries 90.3} & \ulnum{90.1} & {\bfseries 90.1} \\
  \bottomrule
\end{tabular}
\end{table}

\subsection{Multi-Image and Video}
Ovis2.5's proficiency extends to understanding complex multi-modal sequences, which we evaluated on a comprehensive suite of multi-image and video benchmarks. In the multi-image domain, the model's superior cross-image reasoning is evident. On benchmarks like BLINK~\citep{fu2024blink} and MMT-Bench~\citep{ying2024mmt}, it achieves impressive scores of 67.3 and 69.3 respectively, outperforming other prominent models. 

This advantage carries over into video understanding. On a diverse set of video benchmarks—including VideoMME~\citep{fu2025video}, MVBench~\citep{li2024mvbench}, MLVU~\citep{zhou2024mlvu}, and TempCompass~\citep{liu2024tempcompass}—Ovis2.5 maintains a strong and consistent performance across a broader range of tasks. Taken together, these results underscore the model's advanced ability to capture and reason over both spatial and temporal dynamics, establishing its strong competitiveness in sequence-level multimodal understanding.

\begin{table}
  \caption{Performance on multi-image \& video benchmarks. Abbreviations: MMT = MMT-Bench-Val; VM-w/ sub = VideoMME with subtitles; VM-w/o sub = VideoMME without subtitles; MLVU = MLVU (M-Avg); TC = TempCompass}
  \label{tab:und}
  \centering
  \small
  \begin{tabular}{l|S[table-format=2.1] S[table-format=2.1] S[table-format=2.1] S[table-format=2.1] S[table-format=2.1] S[table-format=2.1] S[table-format=2.1]}
    \toprule
    Model & {BLINK} & {MMT} & {VM-w/ sub} & {VM-w/o sub} & {MVBench} & {MLVU} & {TC} \\
    \midrule
    Ovis2-8B                & 55.0 & 66.4 & 71.6 & 68.0 & 68.2 & {\textbf{76.4}} & 69.3 \\
    Qwen2.5-VL-7B           & 56.4 & 63.6 & 71.6 & 65.1 & 69.6 & 70.2 & \ulnum{71.7} \\
    InternVL3-9B            & 58.6 & 65.4 & 68.9 & 66.7 & {\textbf{74.3}} & \ulnum{70.8} & \starnum{71.2} \\
    \midrule
    MiMo-VL-7B-RL-2508      & \starnum{57.8} & \starnum{66.8} & {-} & {\textbf{70.8}} & \starnum{62.0} & \starnum{57.9} & {-} \\
    Keye-VL-8B-Thinking     & 52.0 & \starnum{64.8} & \starnum{64.9} & 67.7 & \starnum{57.6} & \starnum{69.8} & 71.5 \\
    GLM-4.1V-9B-Thinking    & 65.1 & \ulstarnum{67.6} & {\textbf{73.6}} & 68.2 & {-} & {-} & {-} \\
    \midrule
    Ovis2.5-2B              & \ulnum{65.7} & 62.8 & 63.6 & 60.2 & 66.7 & 69.5 & 59.6 \\
    Ovis2.5-9B              & {\textbf{67.3}} & {\textbf{69.3}} & \ulnum{72.8} & \ulnum{68.6} & \ulnum{69.7} & {\textbf{76.4}} & {\textbf{72.6}} \\
    \bottomrule
  \end{tabular}
\end{table}

\section{Conclusion}
\label{sec:conclusion}
In this report, we present Ovis2.5. Building upon the foundational architecture of the Ovis series, this version introduces significant advances in both visual perception and multimodal reasoning. On the perception side, we integrate a NaViT to process images at their native, variable resolutions, preserving both fine details and global structure that are essential for visually dense content such as charts. On the reasoning side, we introduce an advanced ``thinking mode'' paradigm. This capability, cultivated through training on reflection-oriented data, enables the model to perform self-correction and revision on complex tasks, offering a controllable trade-off between latency and accuracy at inference time. Evaluated on a wide spectrum of challenging benchmarks, Ovis2.5 establishes a new state-of-the-art on OpenCompass for its parameter class and delivers leading results across domains including STEM, OCR and chart, visual grounding, and multi-image/video comprehension.

Despite these gains, several promising avenues remain: (i) scaling perception to 4K-level high-resolution images while maintaining accuracy; (ii) handling long-input video with richer temporal reasoning; and (iii) tighter integration of tool use for action-augmented reasoning. We look forward to exploring these directions in our future work.

\section{Core Contributors}
Shiyin Lu, Yang Li, Yu Xia, Yuwei Hu, Shanshan Zhao, Yanqing Ma, Zhichao Wei, Yinglun Li, Lunhao Duan

\section{Contributors}
Jianshan Zhao, Yuxuan Han, Haijun Li, Wanying Chen, Junke Tang, Chengkun Hou, Zhixing Du, Tianli Zhou, Wenjie Zhang, Huping Ding, Jiahe Li, Wen Li, Gui Hu, Yiliang Gu, Siran Yang, Jiamang Wang, Hailong Sun, Yibo Wang, Hui Sun, Jinlong Huang, Yuping He, Shengze Shi, Weihong Zhang, Guodong Zheng, Junpeng Jiang, Sensen Gao, Yi-Feng Wu, Sijia Chen, Yuhui Chen

\section{Project Leaders}
Qing-Guo Chen, Zhao Xu, Weihua Luo, Kaifu Zhang

\bibliography{biblio}

\begin{thebibliography}{50}
\providecommand{\natexlab}[1]{#1}
\providecommand{\url}[1]{\texttt{#1}}
\expandafter\ifx\csname urlstyle\endcsname\relax
  \providecommand{\doi}[1]{doi: #1}\else
  \providecommand{\doi}{doi: \begingroup \urlstyle{rm}\Url}\fi

\bibitem[Bai et~al.(2025)Bai, Chen, Liu, Wang, Ge, Song, Dang, Wang, Wang, Tang, Zhong, Zhu, Yang, Li, Wan, Wang, Ding, Fu, Xu, Ye, Zhang, Xie, Cheng, Zhang, Yang, Xu, and Lin]{Qwen2.5-VL}
Shuai Bai, Keqin Chen, Xuejing Liu, Jialin Wang, Wenbin Ge, Sibo Song, Kai Dang, Peng Wang, Shijie Wang, Jun Tang, Humen Zhong, Yuanzhi Zhu, Mingkun Yang, Zhaohai Li, Jianqiang Wan, Pengfei Wang, Wei Ding, Zheren Fu, Yiheng Xu, Jiabo Ye, Xi~Zhang, Tianbao Xie, Zesen Cheng, Hang Zhang, Zhibo Yang, Haiyang Xu, and Junyang Lin.
\newblock Qwen2.5-vl technical report.
\newblock \emph{arXiv preprint arXiv:2502.13923}, 2025.

\bibitem[Byeon et~al.(2022)Byeon, Park, Kim, Lee, Baek, and Kim]{kakaobrain2022coyo-700m}
Minwoo Byeon, Beomhee Park, Haecheon Kim, Sungjun Lee, Woonhyuk Baek, and Saehoon Kim.
\newblock Coyo-700m: Image-text pair dataset.
\newblock \url{https://github.com/kakaobrain/coyo-dataset}, 2022.

\bibitem[Chen et~al.(2024)Chen, Li, Dong, Zhang, Zang, Chen, Duan, Wang, Qiao, Lin, et~al.]{chen2024mmstar}
Lin Chen, Jinsong Li, Xiaoyi Dong, Pan Zhang, Yuhang Zang, Zehui Chen, Haodong Duan, Jiaqi Wang, Yu~Qiao, Dahua Lin, et~al.
\newblock Are we on the right way for evaluating large vision-language models?
\newblock In \emph{The Thirty-eighth Annual Conference on Neural Information Processing Systems}, 2024.

\bibitem[Dehghani et~al.(2023)Dehghani, Mustafa, Djolonga, Heek, Minderer, Caron, Steiner, Puigcerver, Geirhos, Alabdulmohsin, Oliver, Padlewski, Gritsenko, Lučić, and Houlsby]{dehghani2023patchnpacknavit}
Mostafa Dehghani, Basil Mustafa, Josip Djolonga, Jonathan Heek, Matthias Minderer, Mathilde Caron, Andreas Steiner, Joan Puigcerver, Robert Geirhos, Ibrahim Alabdulmohsin, Avital Oliver, Piotr Padlewski, Alexey Gritsenko, Mario Lučić, and Neil Houlsby.
\newblock Patch n' pack: Navit, a vision transformer for any aspect ratio and resolution, 2023.
\newblock URL \url{https://arxiv.org/abs/2307.06304}.

\bibitem[Duan et~al.(2025)Duan, Fang, Yang, Zhao, Qiao, Li, Agarwal, Chen, Chen, Liu, Ma, Sun, Zhang, Lu, Wong, Wang, Zhou, Li, Fu, Cui, Dong, Zang, Zhang, Wang, Lin, and Chen]{duan2025vlmevalkitopensourcetoolkitevaluating}
Haodong Duan, Xinyu Fang, Junming Yang, Xiangyu Zhao, Yuxuan Qiao, Mo~Li, Amit Agarwal, Zhe Chen, Lin Chen, Yuan Liu, Yubo Ma, Hailong Sun, Yifan Zhang, Shiyin Lu, Tack~Hwa Wong, Weiyun Wang, Peiheng Zhou, Xiaozhe Li, Chaoyou Fu, Junbo Cui, Xiaoyi Dong, Yuhang Zang, Pan Zhang, Jiaqi Wang, Dahua Lin, and Kai Chen.
\newblock Vlmevalkit: An open-source toolkit for evaluating large multi-modality models, 2025.
\newblock URL \url{https://arxiv.org/abs/2407.11691}.

\bibitem[et~al.(2025)]{comanici2025gemini25pushingfrontier}
Gheorghe~Comanici et~al.
\newblock Gemini 2.5: Pushing the frontier with advanced reasoning, multimodality, long context, and next generation agentic capabilities, 2025.
\newblock URL \url{https://arxiv.org/abs/2507.06261}.

\bibitem[Fu et~al.(2025)Fu, Dai, Luo, Li, Ren, Zhang, Wang, Zhou, Shen, Zhang, et~al.]{fu2025video}
Chaoyou Fu, Yuhan Dai, Yongdong Luo, Lei Li, Shuhuai Ren, Renrui Zhang, Zihan Wang, Chenyu Zhou, Yunhang Shen, Mengdan Zhang, et~al.
\newblock Video-mme: The first-ever comprehensive evaluation benchmark of multi-modal llms in video analysis.
\newblock In \emph{Proceedings of the Computer Vision and Pattern Recognition Conference}, pp.\  24108--24118, 2025.

\bibitem[Fu et~al.(2024{\natexlab{a}})Fu, Yang, Kuang, Song, Li, Zhu, Luo, Wang, Lu, Huang, Li, Tang, Shan, Lin, Liu, Wu, Feng, Liu, Huang, Tang, Chen, Jin, Liu, and Bai]{fu2024ocrbenchv2improvedbenchmark}
Ling Fu, Biao Yang, Zhebin Kuang, Jiajun Song, Yuzhe Li, Linghao Zhu, Qidi Luo, Xinyu Wang, Hao Lu, Mingxin Huang, Zhang Li, Guozhi Tang, Bin Shan, Chunhui Lin, Qi~Liu, Binghong Wu, Hao Feng, Hao Liu, Can Huang, Jingqun Tang, Wei Chen, Lianwen Jin, Yuliang Liu, and Xiang Bai.
\newblock Ocrbench v2: An improved benchmark for evaluating large multimodal models on visual text localization and reasoning, 2024{\natexlab{a}}.
\newblock URL \url{https://arxiv.org/abs/2501.00321}.

\bibitem[Fu et~al.(2024{\natexlab{b}})Fu, Hu, Li, Feng, Wang, Lin, Roth, Smith, Ma, and Krishna]{fu2024blink}
Xingyu Fu, Yushi Hu, Bangzheng Li, Yu~Feng, Haoyu Wang, Xudong Lin, Dan Roth, Noah~A Smith, Wei-Chiu Ma, and Ranjay Krishna.
\newblock Blink: Multimodal large language models can see but not perceive.
\newblock In \emph{European Conference on Computer Vision}, pp.\  148--166. Springer, 2024{\natexlab{b}}.

\bibitem[Gadre et~al.(2023)Gadre, Ilharco, Fang, Hayase, Smyrnis, Nguyen, Marten, Wortsman, Ghosh, Zhang, et~al.]{gadre2023datacomp}
Samir~Yitzhak Gadre, Gabriel Ilharco, Alex Fang, Jonathan Hayase, Georgios Smyrnis, Thao Nguyen, Ryan Marten, Mitchell Wortsman, Dhruba Ghosh, Jieyu Zhang, et~al.
\newblock Datacomp: In search of the next generation of multimodal datasets.
\newblock \emph{Advances in Neural Information Processing Systems}, 36:\penalty0 27092--27112, 2023.

\bibitem[Gu et~al.(2022)Gu, Meng, Lu, Hou, Minzhe, Liang, Yao, Huang, Zhang, Jiang, et~al.]{gu2022wukong}
Jiaxi Gu, Xiaojun Meng, Guansong Lu, Lu~Hou, Niu Minzhe, Xiaodan Liang, Lewei Yao, Runhui Huang, Wei Zhang, Xin Jiang, et~al.
\newblock Wukong: A 100 million large-scale chinese cross-modal pre-training benchmark.
\newblock \emph{Advances in Neural Information Processing Systems}, 35:\penalty0 26418--26431, 2022.

\bibitem[Guan et~al.(2024)Guan, Liu, Wu, Xian, Li, Liu, Wang, Chen, Huang, Yacoob, et~al.]{guan2024hallusionbench}
Tianrui Guan, Fuxiao Liu, Xiyang Wu, Ruiqi Xian, Zongxia Li, Xiaoyu Liu, Xijun Wang, Lichang Chen, Furong Huang, Yaser Yacoob, et~al.
\newblock Hallusionbench: an advanced diagnostic suite for entangled language hallucination and visual illusion in large vision-language models.
\newblock In \emph{Proceedings of the IEEE/CVF Conference on Computer Vision and Pattern Recognition}, pp.\  14375--14385, 2024.

\bibitem[Hong et~al.(2025)Hong, Yu, Gu, Wang, Gan, Tang, Cheng, Qi, Ji, Pan, Duan, Wang, Wang, Cheng, He, Su, Yang, Pan, Zeng, Wang, Shi, Pang, Zhang, Yin, Yang, Chen, Xu, Chen, Chen, Chen, Lin, Wang, Chen, Lei, Gong, Pan, Zhang, Zheng, Yang, Zhong, Huang, Zhao, Xue, Tu, Meng, Zhang, Luo, Hao, Li, Jia, Lyu, Huang, Wang, Xue, Wang, An, Du, Shi, Huang, Niu, Wang, Yue, Li, Zhang, Zhang, Du, Hou, Xue, Du, Wang, Zhang, Liu, Xu, Li, Huang, Dong, and Tang]{glmvteam2025glm41vthinkingversatilemultimodalreasoning}
Wenyi Hong, Wenmeng Yu, Xiaotao Gu, Guo Wang, Guobing Gan, Haomiao Tang, Jiale Cheng, Ji~Qi, Junhui Ji, Lihang Pan, Shuaiqi Duan, Weihan Wang, Yan Wang, Yean Cheng, Zehai He, Zhe Su, Zhen Yang, Ziyang Pan, Aohan Zeng, Baoxu Wang, Boyan Shi, Changyu Pang, Chenhui Zhang, Da~Yin, Fan Yang, Guoqing Chen, Jiazheng Xu, Jiali Chen, Jing Chen, Jinhao Chen, Jinghao Lin, Jinjiang Wang, Junjie Chen, Leqi Lei, Letian Gong, Leyi Pan, Mingzhi Zhang, Qinkai Zheng, Sheng Yang, Shi Zhong, Shiyu Huang, Shuyuan Zhao, Siyan Xue, Shangqin Tu, Shengbiao Meng, Tianshu Zhang, Tianwei Luo, Tianxiang Hao, Wenkai Li, Wei Jia, Xin Lyu, Xuancheng Huang, Yanling Wang, Yadong Xue, Yanfeng Wang, Yifan An, Yifan Du, Yiming Shi, Yiheng Huang, Yilin Niu, Yuan Wang, Yuanchang Yue, Yuchen Li, Yutao Zhang, Yuxuan Zhang, Zhanxiao Du, Zhenyu Hou, Zhao Xue, Zhengxiao Du, Zihan Wang, Peng Zhang, Debing Liu, Bin Xu, Juanzi Li, Minlie Huang, Yuxiao Dong, and Jie Tang.
\newblock Glm-4.1v-thinking: Towards versatile multimodal reasoning with scalable reinforcement learning, 2025.
\newblock URL \url{https://arxiv.org/abs/2507.01006}.

\bibitem[Kazemzadeh et~al.(2014{\natexlab{a}})Kazemzadeh, Ordonez, Matten, and Berg]{kazemzadeh-etal-2014-referitgame}
Sahar Kazemzadeh, Vicente Ordonez, Mark Matten, and Tamara Berg.
\newblock {R}efer{I}t{G}ame: Referring to objects in photographs of natural scenes.
\newblock In Alessandro Moschitti, Bo~Pang, and Walter Daelemans (eds.), \emph{Proceedings of the 2014 Conference on Empirical Methods in Natural Language Processing ({EMNLP})}, pp.\  787--798, Doha, Qatar, October 2014{\natexlab{a}}. Association for Computational Linguistics.
\newblock \doi{10.3115/v1/D14-1086}.
\newblock URL \url{https://aclanthology.org/D14-1086}.

\bibitem[Kazemzadeh et~al.(2014{\natexlab{b}})Kazemzadeh, Ordonez, Matten, and Berg]{kazemzadeh2014referitgame}
Sahar Kazemzadeh, Vicente Ordonez, Mark Matten, and Tamara Berg.
\newblock Referitgame: Referring to objects in photographs of natural scenes.
\newblock In \emph{Proceedings of the 2014 conference on empirical methods in natural language processing (EMNLP)}, pp.\  787--798, 2014{\natexlab{b}}.

\bibitem[Kembhavi et~al.(2016)Kembhavi, Salvato, Kolve, Seo, Hajishirzi, and Farhadi]{kembhavi2016diagram}
Aniruddha Kembhavi, Mike Salvato, Eric Kolve, Minjoon Seo, Hannaneh Hajishirzi, and Ali Farhadi.
\newblock A diagram is worth a dozen images.
\newblock In \emph{European conference on computer vision}, pp.\  235--251. Springer, 2016.

\bibitem[Kirillov et~al.(2023)Kirillov, Mintun, Ravi, Mao, Rolland, Gustafson, Xiao, Whitehead, Berg, Lo, et~al.]{kirillov2023segment}
Alexander Kirillov, Eric Mintun, Nikhila Ravi, Hanzi Mao, Chloe Rolland, Laura Gustafson, Tete Xiao, Spencer Whitehead, Alexander~C Berg, Wan-Yen Lo, et~al.
\newblock Segment anything.
\newblock In \emph{Proceedings of the IEEE/CVF international conference on computer vision}, pp.\  4015--4026, 2023.

\bibitem[Li et~al.(2024)Li, Wang, He, Li, Wang, Liu, Wang, Xu, Chen, Luo, et~al.]{li2024mvbench}
Kunchang Li, Yali Wang, Yinan He, Yizhuo Li, Yi~Wang, Yi~Liu, Zun Wang, Jilan Xu, Guo Chen, Ping Luo, et~al.
\newblock Mvbench: A comprehensive multi-modal video understanding benchmark.
\newblock In \emph{Proceedings of the IEEE/CVF Conference on Computer Vision and Pattern Recognition}, pp.\  22195--22206, 2024.

\bibitem[Liu et~al.(2024{\natexlab{a}})Liu, Li, Li, Li, Zhang, Shen, and Lee]{liu2024llavanext}
Haotian Liu, Chunyuan Li, Yuheng Li, Bo~Li, Yuanhan Zhang, Sheng Shen, and Yong~Jae Lee.
\newblock Llava-next: Improved reasoning, ocr, and world knowledge, January 2024{\natexlab{a}}.
\newblock URL \url{https://llava-vl.github.io/blog/2024-01-30-llava-next/}.

\bibitem[Liu et~al.(2024{\natexlab{b}})Liu, Duan, Zhang, Li, Zhang, Zhao, Yuan, Wang, He, Liu, et~al.]{liu2024mmbench}
Yuan Liu, Haodong Duan, Yuanhan Zhang, Bo~Li, Songyang Zhang, Wangbo Zhao, Yike Yuan, Jiaqi Wang, Conghui He, Ziwei Liu, et~al.
\newblock Mmbench: Is your multi-modal model an all-around player?
\newblock In \emph{European conference on computer vision}, pp.\  216--233. Springer, 2024{\natexlab{b}}.

\bibitem[Liu et~al.(2024{\natexlab{c}})Liu, Li, Liu, Wang, Ren, Li, Chen, Sun, and Hou]{liu2024tempcompass}
Yuanxin Liu, Shicheng Li, Yi~Liu, Yuxiang Wang, Shuhuai Ren, Lei Li, Sishuo Chen, Xu~Sun, and Lu~Hou.
\newblock Tempcompass: Do video llms really understand videos?
\newblock \emph{arXiv preprint arXiv:2403.00476}, 2024{\natexlab{c}}.

\bibitem[Liu et~al.(2024{\natexlab{d}})Liu, Li, Huang, Yang, Yu, Li, Yin, Liu, Jin, and Bai]{liu2024ocrbench}
Yuliang Liu, Zhang Li, Mingxin Huang, Biao Yang, Wenwen Yu, Chunyuan Li, Xu-Cheng Yin, Cheng-Lin Liu, Lianwen Jin, and Xiang Bai.
\newblock Ocrbench: on the hidden mystery of ocr in large multimodal models.
\newblock \emph{Science China Information Sciences}, 67\penalty0 (12):\penalty0 220102, 2024{\natexlab{d}}.

\bibitem[Lu et~al.(2024{\natexlab{a}})Lu, Bansal, Xia, Liu, Li, Hajishirzi, Cheng, Chang, Galley, and Gao]{lu2024mathvista}
Pan Lu, Hritik Bansal, Tony Xia, Jiacheng Liu, Chunyuan Li, Hannaneh Hajishirzi, Hao Cheng, Kai-Wei Chang, Michel Galley, and Jianfeng Gao.
\newblock Mathvista: Evaluating mathematical reasoning of foundation models in visual contexts.
\newblock In \emph{The Twelfth International Conference on Learning Representations}, 2024{\natexlab{a}}.

\bibitem[Lu et~al.(2024{\natexlab{b}})Lu, Li, Chen, Xu, Luo, Zhang, and Ye]{lu2024ovis}
Shiyin Lu, Yang Li, Qing-Guo Chen, Zhao Xu, Weihua Luo, Kaifu Zhang, and Han-Jia Ye.
\newblock Ovis: Structural embedding alignment for multimodal large language model.
\newblock \emph{arXiv:2405.20797}, 2024{\natexlab{b}}.

\bibitem[Mao et~al.(2016)Mao, Huang, Toshev, Camburu, Yuille, and Murphy]{mao2016generation}
Junhua Mao, Jonathan Huang, Alexander Toshev, Oana Camburu, Alan~L Yuille, and Kevin Murphy.
\newblock Generation and comprehension of unambiguous object descriptions.
\newblock In \emph{Proceedings of the IEEE conference on computer vision and pattern recognition}, pp.\  11--20, 2016.

\bibitem[Masry et~al.(2022)Masry, Long, Tan, Joty, and Hoque]{masry-etal-2022-chartqa}
Ahmed Masry, Do~Long, Jia~Qing Tan, Shafiq Joty, and Enamul Hoque.
\newblock {C}hart{QA}: A benchmark for question answering about charts with visual and logical reasoning.
\newblock In \emph{Findings of the Association for Computational Linguistics: ACL 2022}, pp.\  2263--2279, Dublin, Ireland, May 2022. Association for Computational Linguistics.
\newblock \doi{10.18653/v1/2022.findings-acl.177}.
\newblock URL \url{https://aclanthology.org/2022.findings-acl.177}.

\bibitem[Masry et~al.(2025)Masry, Islam, Ahmed, Bajaj, Kabir, Kartha, Laskar, Rahman, Rahman, Shahmohammadi, Thakkar, Parvez, Hoque, and Joty]{masry2025chartqaprodiversechallengingbenchmark}
Ahmed Masry, Mohammed~Saidul Islam, Mahir Ahmed, Aayush Bajaj, Firoz Kabir, Aaryaman Kartha, Md~Tahmid~Rahman Laskar, Mizanur Rahman, Shadikur Rahman, Mehrad Shahmohammadi, Megh Thakkar, Md~Rizwan Parvez, Enamul Hoque, and Shafiq Joty.
\newblock Chartqapro: A more diverse and challenging benchmark for chart question answering, 2025.
\newblock URL \url{https://arxiv.org/abs/2504.05506}.

\bibitem[Mathew et~al.(2021)Mathew, Karatzas, and Jawahar]{mathew2021docvqa}
Minesh Mathew, Dimosthenis Karatzas, and CV~Jawahar.
\newblock Docvqa: A dataset for vqa on document images.
\newblock In \emph{Proceedings of the IEEE/CVF winter conference on applications of computer vision}, pp.\  2200--2209, 2021.

\bibitem[Qiao et~al.(2024)Qiao, Tan, Dong, Wu, Sun, Song, Gongque, Lei, Wei, Zhang, Qiao, Zhang, Zong, Xu, Diao, Bao, Li, and Zhang]{qiao2024we}
Runqi Qiao, Qiuna Tan, Guanting Dong, Minhui Wu, Chong Sun, Xiaoshuai Song, Zhuoma Gongque, Shanglin Lei, Zhe Wei, Miaoxuan Zhang, Runfeng Qiao, Yifan Zhang, Xiao Zong, Yida Xu, Muxi Diao, Zhimin Bao, Chen Li, and Honggang Zhang.
\newblock We-math: Does your large multimodal model achieve human-like mathematical reasoning?
\newblock \emph{CoRR}, abs/2407.01284, 2024.
\newblock \doi{10.48550/ARXIV.2407.01284}.
\newblock URL \url{https://doi.org/10.48550/arXiv.2407.01284}.

\bibitem[Rafailov et~al.(2023)Rafailov, Sharma, Mitchell, Manning, Ermon, and Finn]{rafailov2023direct}
Rafael Rafailov, Archit Sharma, Eric Mitchell, Christopher~D Manning, Stefano Ermon, and Chelsea Finn.
\newblock Direct preference optimization: Your language model is secretly a reward model.
\newblock \emph{Advances in neural information processing systems}, 36:\penalty0 53728--53741, 2023.

\bibitem[Schuhmann et~al.(2022)Schuhmann, Beaumont, Vencu, Gordon, Wightman, Cherti, Coombes, Katta, Mullis, Wortsman, et~al.]{schuhmann2022laion}
Christoph Schuhmann, Romain Beaumont, Richard Vencu, Cade Gordon, Ross Wightman, Mehdi Cherti, Theo Coombes, Aarush Katta, Clayton Mullis, Mitchell Wortsman, et~al.
\newblock Laion-5b: An open large-scale dataset for training next generation image-text models.
\newblock \emph{Advances in neural information processing systems}, 35:\penalty0 25278--25294, 2022.

\bibitem[Shao et~al.(2024)Shao, Wang, Zhu, Xu, Song, Bi, Zhang, Zhang, Li, Wu, et~al.]{shao2024deepseekmath}
Zhihong Shao, Peiyi Wang, Qihao Zhu, Runxin Xu, Junxiao Song, Xiao Bi, Haowei Zhang, Mingchuan Zhang, YK~Li, Yang Wu, et~al.
\newblock Deepseekmath: Pushing the limits of mathematical reasoning in open language models.
\newblock \emph{arXiv preprint arXiv:2402.03300}, 2024.

\bibitem[Shoeybi et~al.(2019)Shoeybi, Patwary, Puri, LeGresley, Casper, and Catanzaro]{shoeybi2019megatron}
Mohammad Shoeybi, Mostofa Patwary, Raul Puri, Patrick LeGresley, Jared Casper, and Bryan Catanzaro.
\newblock Megatron-lm: Training multi-billion parameter language models using model parallelism.
\newblock \emph{arXiv preprint arXiv:1909.08053}, 2019.

\bibitem[Singh et~al.(2019)Singh, Natarajan, Shah, Jiang, Chen, Batra, Parikh, and Rohrbach]{singh2019towards}
Amanpreet Singh, Vivek Natarajan, Meet Shah, Yu~Jiang, Xinlei Chen, Dhruv Batra, Devi Parikh, and Marcus Rohrbach.
\newblock Towards vqa models that can read.
\newblock In \emph{Proceedings of the IEEE/CVF conference on computer vision and pattern recognition}, pp.\  8317--8326, 2019.

\bibitem[Su et~al.(2024)Su, Ahmed, Lu, Pan, Bo, and Liu]{su2024roformer}
Jianlin Su, Murtadha Ahmed, Yu~Lu, Shengfeng Pan, Wen Bo, and Yunfeng Liu.
\newblock Roformer: Enhanced transformer with rotary position embedding.
\newblock \emph{Neurocomputing}, 568:\penalty0 127063, 2024.

\bibitem[Tschannen et~al.(2025)Tschannen, Gritsenko, Wang, Naeem, Alabdulmohsin, Parthasarathy, Evans, Beyer, Xia, Mustafa, Hénaff, Harmsen, Steiner, and Zhai]{tschannen2025siglip2multilingualvisionlanguage}
Michael Tschannen, Alexey Gritsenko, Xiao Wang, Muhammad~Ferjad Naeem, Ibrahim Alabdulmohsin, Nikhil Parthasarathy, Talfan Evans, Lucas Beyer, Ye~Xia, Basil Mustafa, Olivier Hénaff, Jeremiah Harmsen, Andreas Steiner, and Xiaohua Zhai.
\newblock Siglip 2: Multilingual vision-language encoders with improved semantic understanding, localization, and dense features, 2025.
\newblock URL \url{https://arxiv.org/abs/2502.14786}.

\bibitem[Wang et~al.(2024)Wang, Pan, Shi, Lu, Ren, Zhou, Zhan, and Li]{wang2024measuring}
Ke~Wang, Junting Pan, Weikang Shi, Zimu Lu, Houxing Ren, Aojun Zhou, Mingjie Zhan, and Hongsheng Li.
\newblock Measuring multimodal mathematical reasoning with math-vision dataset.
\newblock \emph{Advances in Neural Information Processing Systems}, 37:\penalty0 95095--95169, 2024.

\bibitem[Xiao et~al.(2024)Xiao, Sun, Liu, and Wang]{xiao2024logicvistamultimodalllmlogical}
Yijia Xiao, Edward Sun, Tianyu Liu, and Wei Wang.
\newblock Logicvista: Multimodal llm logical reasoning benchmark in visual contexts, 2024.
\newblock URL \url{https://arxiv.org/abs/2407.04973}.

\bibitem[Yang et~al.(2024)Yang, Yang, Zhang, Hui, Zheng, Yu, Li, Liu, Huang, Wei, Lin, Yang, Tu, Zhang, Yang, Yang, Zhou, Lin, Dang, Lu, Bao, Yang, Yu, Li, Xue, Zhang, Zhu, Men, Lin, Li, Xia, Ren, Ren, Fan, Su, Zhang, Wan, Liu, Cui, Zhang, and Qiu]{an2024qwen2.5}
An~Yang, Baosong Yang, Beichen Zhang, Binyuan Hui, Bo~Zheng, Bowen Yu, Chengyuan Li, Dayiheng Liu, Fei Huang, Haoran Wei, Huan Lin, Jian Yang, Jianhong Tu, Jianwei Zhang, Jianxin Yang, Jiaxi Yang, Jingren Zhou, Junyang Lin, Kai Dang, Keming Lu, Keqin Bao, Kexin Yang, Le~Yu, Mei Li, Mingfeng Xue, Pei Zhang, Qin Zhu, Rui Men, Runji Lin, Tianhao Li, Tingyu Xia, Xingzhang Ren, Xuancheng Ren, Yang Fan, Yang Su, Yichang Zhang, Yu~Wan, Yuqiong Liu, Zeyu Cui, Zhenru Zhang, and Zihan Qiu.
\newblock Qwen2.5 technical report.
\newblock \emph{arXiv preprint arXiv:2412.15115}, 2024.

\bibitem[Yang et~al.(2025{\natexlab{a}})Yang, Li, Yang, Zhang, Hui, Zheng, Yu, Gao, Huang, Lv, Zheng, Liu, Zhou, Huang, Hu, Ge, Wei, Lin, Tang, Yang, Tu, Zhang, Yang, Yang, Zhou, Zhou, Lin, Dang, Bao, Yang, Yu, Deng, Li, Xue, Li, Zhang, Wang, Zhu, Men, Gao, Liu, Luo, Li, Tang, Yin, Ren, Wang, Zhang, Ren, Fan, Su, Zhang, Zhang, Wan, Liu, Wang, Cui, Zhang, Zhou, and Qiu]{qwen3}
An~Yang, Anfeng Li, Baosong Yang, Beichen Zhang, Binyuan Hui, Bo~Zheng, Bowen Yu, Chang Gao, Chengen Huang, Chenxu Lv, Chujie Zheng, Dayiheng Liu, Fan Zhou, Fei Huang, Feng Hu, Hao Ge, Haoran Wei, Huan Lin, Jialong Tang, Jian Yang, Jianhong Tu, Jianwei Zhang, Jianxin Yang, Jiaxi Yang, Jing Zhou, Jingren Zhou, Junyang Lin, Kai Dang, Keqin Bao, Kexin Yang, Le~Yu, Lianghao Deng, Mei Li, Mingfeng Xue, Mingze Li, Pei Zhang, Peng Wang, Qin Zhu, Rui Men, Ruize Gao, Shixuan Liu, Shuang Luo, Tianhao Li, Tianyi Tang, Wenbiao Yin, Xingzhang Ren, Xinyu Wang, Xinyu Zhang, Xuancheng Ren, Yang Fan, Yang Su, Yichang Zhang, Yinger Zhang, Yu~Wan, Yuqiong Liu, Zekun Wang, Zeyu Cui, Zhenru Zhang, Zhipeng Zhou, and Zihan Qiu.
\newblock Qwen3 technical report.
\newblock \emph{arXiv preprint arXiv:2505.09388}, 2025{\natexlab{a}}.

\bibitem[Yang et~al.(2025{\natexlab{b}})Yang, Wen, Liu, Chu, Song, Rao, Yi, Li, Zang, Yang, Zhou, Peng, Ding, Huang, Cao, Chen, Hua, Ouyang, Chen, Jiang, Tang, Gai, Zhang, Mao, Huang, Zhang, Gao, Chen, Yuan, Wu, Hu, Lu, Zhou, Zhang, Yang, Chen, Wu, Li, Ling, Li, Ma, Xu, Gao, Li, Guo, Wang, Ren, Wei, Wang, Hu, Wang, Yu, Luo, Li, Liang, Hu, Lu, Yang, and Zhang]{kwaikeyeteam2025kwaikeyevltechnicalreport}
Biao Yang, Bin Wen, Changyi Liu, Chenglong Chu, Chengru Song, Chongling Rao, Chuan Yi, Da~Li, Dunju Zang, Fan Yang, Guorui Zhou, Hao Peng, Haojie Ding, Jiaming Huang, Jiangxia Cao, Jiankang Chen, Jingyun Hua, Jin Ouyang, Kaibing Chen, Kaiyu Jiang, Kaiyu Tang, Kun Gai, Shengnan Zhang, Siyang Mao, Sui Huang, Tianke Zhang, Tingting Gao, Wei Chen, Wei Yuan, Xiangyu Wu, Xiao Hu, Xingyu Lu, Yang Zhou, Yi-Fan Zhang, Yiping Yang, Yulong Chen, Zhenhua Wu, Zhenyu Li, Zhixin Ling, Ziming Li, Dehua Ma, Di~Xu, Haixuan Gao, Hang Li, Jiawei Guo, Jing Wang, Lejian Ren, Muhao Wei, Qianqian Wang, Qigen Hu, Shiyao Wang, Tao Yu, Xinchen Luo, Yan Li, Yiming Liang, Yuhang Hu, Zeyi Lu, Zhuoran Yang, and Zixing Zhang.
\newblock Kwai keye-vl technical report, 2025{\natexlab{b}}.
\newblock URL \url{https://arxiv.org/abs/2507.01949}.

\bibitem[Ying et~al.(2024)Ying, Meng, Wang, Li, Lin, Yang, Zhang, Zhang, Lin, Liu, et~al.]{ying2024mmt}
Kaining Ying, Fanqing Meng, Jin Wang, Zhiqian Li, Han Lin, Yue Yang, Hao Zhang, Wenbo Zhang, Yuqi Lin, Shuo Liu, et~al.
\newblock Mmt-bench: A comprehensive multimodal benchmark for evaluating large vision-language models towards multitask agi.
\newblock \emph{arXiv preprint arXiv:2404.16006}, 2024.

\bibitem[Yu et~al.(2023)Yu, Yang, Li, Wang, Lin, Liu, Wang, and Wang]{yu2023mmvet}
Weihao Yu, Zhengyuan Yang, Linjie Li, Jianfeng Wang, Kevin Lin, Zicheng Liu, Xinchao Wang, and Lijuan Wang.
\newblock Mm-vet: Evaluating large multimodal models for integrated capabilities.
\newblock \emph{arXiv preprint arXiv:2308.02490}, 2023.

\bibitem[Yue et~al.(2024{\natexlab{a}})Yue, Ni, Zhang, Zheng, Liu, Zhang, Stevens, Jiang, Ren, Sun, et~al.]{yue2024mmmu}
Xiang Yue, Yuansheng Ni, Kai Zhang, Tianyu Zheng, Ruoqi Liu, Ge~Zhang, Samuel Stevens, Dongfu Jiang, Weiming Ren, Yuxuan Sun, et~al.
\newblock Mmmu: A massive multi-discipline multimodal understanding and reasoning benchmark for expert agi.
\newblock In \emph{Proceedings of the IEEE/CVF Conference on Computer Vision and Pattern Recognition}, pp.\  9556--9567, 2024{\natexlab{a}}.

\bibitem[Yue et~al.(2024{\natexlab{b}})Yue, Zheng, Ni, Wang, Zhang, Tong, Sun, Yu, Zhang, Sun, et~al.]{yue2024mmmupro}
Xiang Yue, Tianyu Zheng, Yuansheng Ni, Yubo Wang, Kai Zhang, Shengbang Tong, Yuxuan Sun, Botao Yu, Ge~Zhang, Huan Sun, et~al.
\newblock Mmmu-pro: A more robust multi-discipline multimodal understanding benchmark.
\newblock \emph{arXiv preprint arXiv:2409.02813}, 2024{\natexlab{b}}.

\bibitem[Yue et~al.(2025)Yue, Lin, Song, Wang, Ren, Gu, Li, Li, Zhao, Li, Bao, Tian, Zhang, Wang, Zhu, Cici, He, Ye, Shen, Zhang, Jiang, Zheng, Song, Luo, Yu, Wang, Tian, Tu, Yan, Huang, Wang, Xu, Song, Zhang, Yong, Zhang, Deng, Yang, Ma, Lv, Zhuang, Liu, Deng, Liu, Chen, Yu, Liu, Wang, Ma, Wang, Wang, Chen, Zhu, Zhou, Zhou, Fang, Shi, Dong, Xiao, Xu, Liu, Xu, Qu, Zhao, Lv, Wang, Zhang, Zhang, Zhang, Ma, Liu, Cai, and Xia]{coreteam2025mimovltechnicalreport}
Zihao Yue, Zhenru Lin, Yifan Song, Weikun Wang, Shuhuai Ren, Shuhao Gu, Shicheng Li, Peidian Li, Liang Zhao, Lei Li, Kainan Bao, Hao Tian, Hailin Zhang, Gang Wang, Dawei Zhu, Cici, Chenhong He, Bowen Ye, Bowen Shen, Zihan Zhang, Zihan Jiang, Zhixian Zheng, Zhichao Song, Zhenbo Luo, Yue Yu, Yudong Wang, Yuanyuan Tian, Yu~Tu, Yihan Yan, Yi~Huang, Xu~Wang, Xinzhe Xu, Xingchen Song, Xing Zhang, Xing Yong, Xin Zhang, Xiangwei Deng, Wenyu Yang, Wenhan Ma, Weiwei Lv, Weiji Zhuang, Wei Liu, Sirui Deng, Shuo Liu, Shimao Chen, Shihua Yu, Shaohui Liu, Shande Wang, Rui Ma, Qiantong Wang, Peng Wang, Nuo Chen, Menghang Zhu, Kangyang Zhou, Kang Zhou, Kai Fang, Jun Shi, Jinhao Dong, Jiebao Xiao, Jiaming Xu, Huaqiu Liu, Hongshen Xu, Heng Qu, Haochen Zhao, Hanglong Lv, Guoan Wang, Duo Zhang, Dong Zhang, Di~Zhang, Chong Ma, Chang Liu, Can Cai, and Bingquan Xia.
\newblock Mimo-vl technical report, 2025.
\newblock URL \url{https://arxiv.org/abs/2506.03569}.

\bibitem[Zhang et~al.(2024)Zhang, Jiang, Zhang, Lin, Guo, Qiu, Zhou, Lu, Chang, Qiao, et~al.]{zhang2024mathverse}
Renrui Zhang, Dongzhi Jiang, Yichi Zhang, Haokun Lin, Ziyu Guo, Pengshuo Qiu, Aojun Zhou, Pan Lu, Kai-Wei Chang, Yu~Qiao, et~al.
\newblock Mathverse: Does your multi-modal llm truly see the diagrams in visual math problems?
\newblock In \emph{European Conference on Computer Vision}, pp.\  169--186. Springer, 2024.

\bibitem[Zhou et~al.(2024)Zhou, Shu, Zhao, Wu, Xiao, Yang, Xiong, Zhang, Huang, and Liu]{zhou2024mlvu}
Junjie Zhou, Yan Shu, Bo~Zhao, Boya Wu, Shitao Xiao, Xi~Yang, Yongping Xiong, Bo~Zhang, Tiejun Huang, and Zheng Liu.
\newblock Mlvu: A comprehensive benchmark for multi-task long video understanding.
\newblock \emph{arXiv e-prints}, pp.\  arXiv--2406, 2024.

\bibitem[Zhu et~al.(2025)Zhu, Wang, Chen, Liu, Ye, Gu, Tian, Duan, Su, Shao, Gao, Cui, Wang, Cao, Liu, Wei, Zhang, Wang, Xu, Li, Wang, Deng, Li, He, Jiang, Luo, Wang, He, Shi, Zhang, Shao, He, Xiong, Qu, Sun, Jiao, Lv, Wu, Zhang, Deng, Ge, Chen, Wang, Dou, Lu, Zhu, Lu, Lin, Qiao, Dai, and Wang]{zhu2025internvl3exploringadvancedtraining}
Jinguo Zhu, Weiyun Wang, Zhe Chen, Zhaoyang Liu, Shenglong Ye, Lixin Gu, Hao Tian, Yuchen Duan, Weijie Su, Jie Shao, Zhangwei Gao, Erfei Cui, Xuehui Wang, Yue Cao, Yangzhou Liu, Xingguang Wei, Hongjie Zhang, Haomin Wang, Weiye Xu, Hao Li, Jiahao Wang, Nianchen Deng, Songze Li, Yinan He, Tan Jiang, Jiapeng Luo, Yi~Wang, Conghui He, Botian Shi, Xingcheng Zhang, Wenqi Shao, Junjun He, Yingtong Xiong, Wenwen Qu, Peng Sun, Penglong Jiao, Han Lv, Lijun Wu, Kaipeng Zhang, Huipeng Deng, Jiaye Ge, Kai Chen, Limin Wang, Min Dou, Lewei Lu, Xizhou Zhu, Tong Lu, Dahua Lin, Yu~Qiao, Jifeng Dai, and Wenhai Wang.
\newblock Internvl3: Exploring advanced training and test-time recipes for open-source multimodal models, 2025.
\newblock URL \url{https://arxiv.org/abs/2504.10479}.

\bibitem[Zou et~al.(2024)Zou, Guo, Yang, Zhang, Hu, and Zhang]{zou2024dynamic}
Chengke Zou, Xingang Guo, Rui Yang, Junyu Zhang, Bin Hu, and Huan Zhang.
\newblock Dynamath: A dynamic visual benchmark for evaluating mathematical reasoning robustness of vision language models, 2024.

\end{thebibliography}
\bibliographystyle{colm2024_conference}

\newpage
\appendix

\section{Qualitative Examples}

In this section, we present a variety of qualitative examples showcasing the capabilities and strengths of the Ovis2.5 model through responses to diverse prompts.

\subsection{World Knowledge}

\begin{figure}[H]
  \centering
  \noindent\color{black}\rule{\linewidth}{0.5pt}
  \vspace{3pt}

  \includegraphics[width=0.9\linewidth,height=0.45\textheight,keepaspectratio]{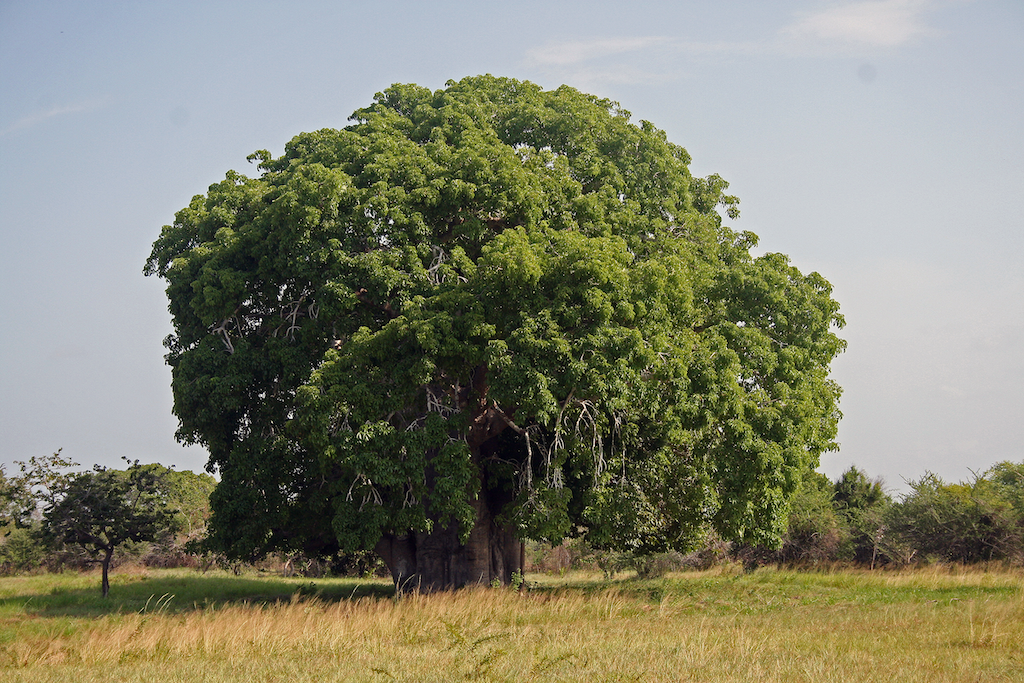}

  \vspace{6pt}
  \begin{minipage}{0.95\linewidth}
    \textbf{Prompts:} What plant is this?

    \medskip
    \textbf{Model Responses:} 

    \texttt{<think>}...\texttt{</think>} This plant is a \textbf{baobab tree} (scientific name \textit{Adansonia digitata}). Baobab trees are iconic in many African savannas, known for their massive trunks, which store water, and their distinctive rounded canopies. They play vital ecological roles and hold cultural significance in regions where they grow.
  \end{minipage}

  \vspace{6pt}
  \noindent\color{black}\rule{\linewidth}{0.5pt}
  \captionsetup{font=small}
  \caption{An example illustrating the model's ability to identify plant species.}
\end{figure}

\begin{figure}[H]
  \centering
  \noindent\color{black}\rule{\linewidth}{0.5pt}
  \vspace{3pt}

  \includegraphics[width=0.9\linewidth,height=0.45\textheight,keepaspectratio]{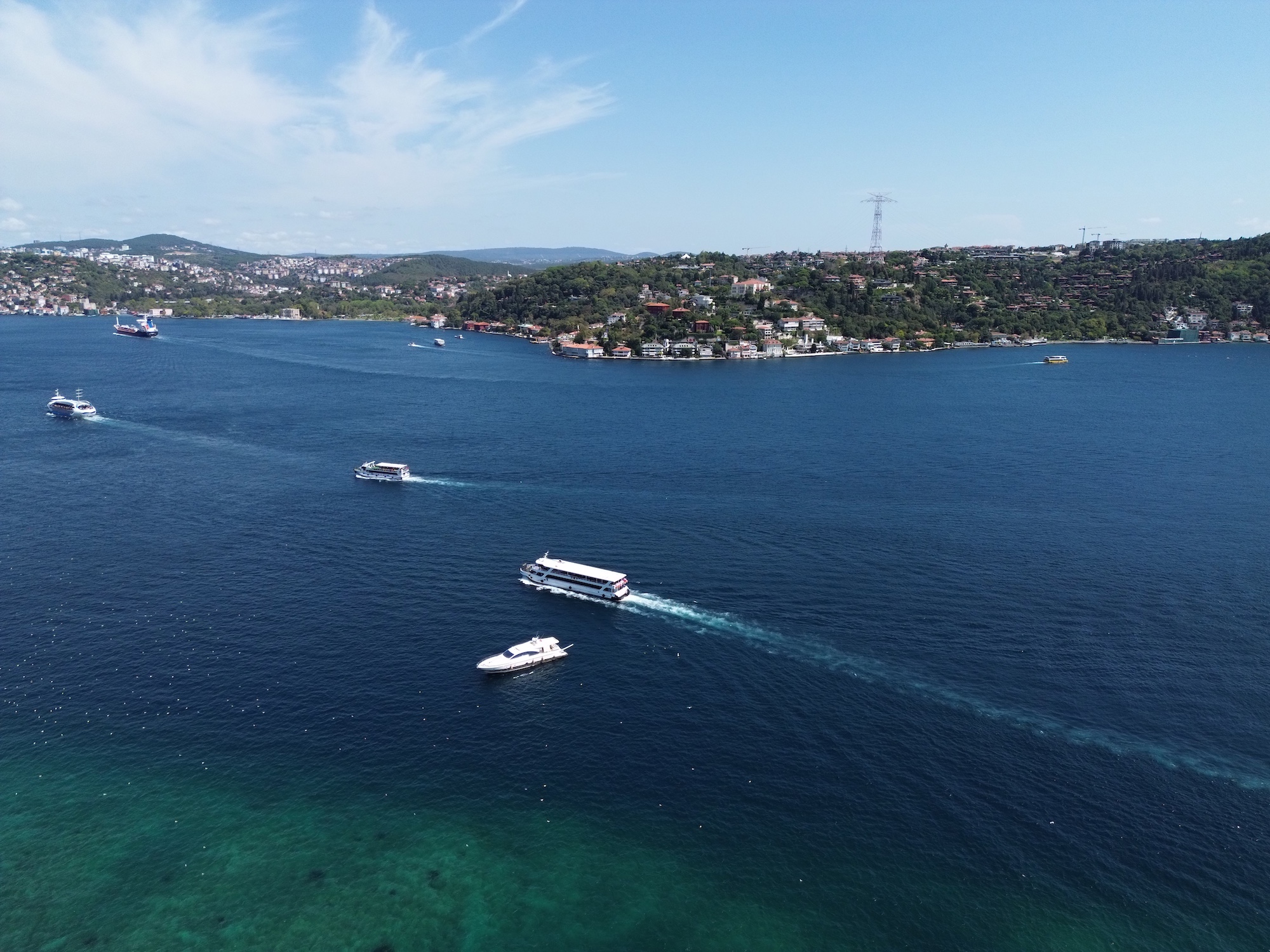}

  \vspace{6pt}
  \begin{minipage}{0.95\linewidth}
    \textbf{Prompts:} Where could this possibly be?

    \medskip
    \textbf{Model Responses:} 

    \texttt{<think>}...\texttt{</think>} This image likely depicts a coastal area in \textbf{Istanbul, Turkey}, specifically near the Bosphorus Strait. Here's why:

\begin{enumerate}
  \item \textbf{Waterway \& Boats:} The wide, deep blue water with ferry-like vessels (common for Istanbul's Bosphorus crossings) and smaller boats matches the bustling maritime activity of the Bosphorus.
  \item \textbf{Landscape \& Architecture:} The hilly terrain with dense residential buildings (typical of Istanbul's shoreline) and the mix of greenery and urban development aligns with the city's geography.
  \item \textbf{Context of Water Traffic:} Istanbul is renowned for its busy waterways connecting the Black Sea and the Sea of Marmara via the Bosphorus, with frequent ferries and tour boats---exactly what the image shows.
\end{enumerate}

While other coastal regions (e.g., parts of Greece or Mediterranean cities) could share some traits, Istanbul's unique combination of water traffic, topography, and architectural style makes this the most plausible location.
  \end{minipage}

  \vspace{6pt}
  \noindent\color{black}\rule{\linewidth}{0.5pt}
  \captionsetup{font=small}
  \caption{An example illustrating the model's ability to recognize geographic locations.}
\end{figure}

\subsection{Math Problem Solving}

\begin{figure}[H]
  \centering
  \noindent\color{black}\rule{\linewidth}{0.5pt}
  \vspace{3pt}

  \includegraphics[width=0.45\linewidth,height=0.45\textheight,keepaspectratio]{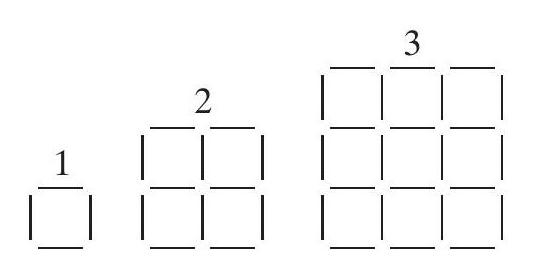}

  \vspace{6pt}
  \begin{minipage}{0.95\linewidth}
    \textbf{Prompts:} Belinda is making patterns with toothpicks according to the schema of the figure. How many toothpicks does Belinda add to the 30th pattern to make the 31 st?

    \medskip
    \textbf{Model Responses:} 

    \texttt{<think>} Okay, let's try to figure out this problem step by step. First, we need to understand how the number of toothpicks changes from one pattern to the next. Let's start by counting the toothpicks in each pattern.

    First, let's recall that these are square patterns. The first pattern (n=1) is a single square. A square has 4 sides, so that's 4 toothpicks. Wait, but let's check the second pattern. The second pattern (n=2) is a 2x2 grid of squares. How many toothpicks does that have? Let's count: horizontally, there are 3 rows of toothpicks, each with 2 segments, so 3*2 = 6. Vertically, there are 3 columns of toothpicks, each with 2 segments, so 3*2 = 6. Total is 12. Wait, or maybe another way: for an n x n grid of squares, the number of horizontal toothpicks is (n+1)*n, and vertical toothpicks is also (n+1)*n. So total toothpicks would be 2*(n+1)*n. Let's check:
    
    For n=1: 2*(2)*1 = 4. Correct.
    
    For n=2: 2*(3)*2 = 12. Let's count: the second pattern has 2 rows and 2 columns of squares, so horizontally, there are 3 lines (top, middle, bottom) each with 2 toothpicks, so 3*2=6. Vertically, 3 lines (left, middle, right) each with 2 toothpicks, so 3*2=6. Total 12. Correct.

    \vspace{0.25\baselineskip}
    \emph{\textcolor{gray}{[Remaining reasoning omitted for brevity.]}} \texttt{</think>}
    \vspace{0.5\baselineskip}

    To determine how many toothpicks Belinda adds to the 30th pattern to make the 31st, we first analyze the pattern of toothpick counts.

    \subsection*{Step 1: Identify the formula for the \( n \)-th pattern}
    The patterns are square grids of size \( n \times n \). For an \( n \times n \) grid of squares:
    \begin{itemize}
        \item \textbf{Horizontal toothpicks}: There are \( n+1 \) horizontal lines, each with \( n \) segments. Thus, total horizontal toothpicks \(= n(n+1) \).
        \item \textbf{Vertical toothpicks}: Similarly, there are \( n+1 \) vertical lines, each with \( n \) segments. Thus, total vertical toothpicks \(= n(n+1) \).
    \end{itemize}
    
    Adding these, the total number of toothpicks in the \( n \)-th pattern is:
    \[
    T(n) = 2n(n+1)
    \]
    
    \subsection*{Step 2: Find the difference between consecutive patterns}
    To find how many toothpicks are added from the \( n \)-th to the \( (n+1) \)-th pattern, calculate \( T(n+1) - T(n) \):
    \[
    \begin{aligned}
    T(n+1) - T(n) &= 2(n+1)(n+2) - 2n(n+1) \\
    &= 2(n+1)\left[(n+2) - n\right] \\
    &= 2(n+1)(2) \\
    &= 4(n+1)
    \end{aligned}
    \]
    
    \subsection*{Step 3: Apply the formula to the 30th to 31st pattern}
    For the transition from the 30th to the 31st pattern, \( n = 30 \). Substitute \( n = 30 \) into the difference formula:
    \[
    T(31) - T(30) = 4(30 + 1) = 4 \times 31 = 124
    \]
    
    Thus, Belinda adds \(\mathbf{124}\) toothpicks to the 30th pattern to make the 31st.
    
  \end{minipage}

  \vspace{6pt}
  \noindent\color{black}\rule{\linewidth}{0.5pt}
  \captionsetup{font=small}
  \caption{An example illustrating the model's ability to solve combinatorial pattern problems.}
\end{figure}

\begin{figure}[H]
  \centering
  \noindent\color{black}\rule{\linewidth}{0.5pt}
  \vspace{3pt}

  \includegraphics[width=0.25\linewidth,height=0.45\textheight,keepaspectratio]{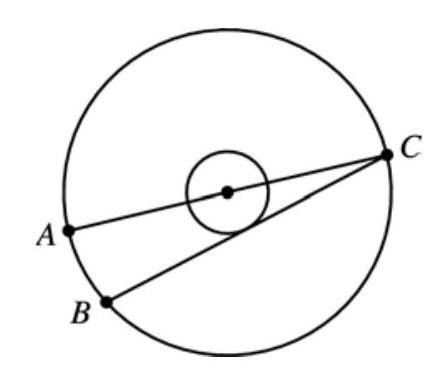}

  \vspace{6pt}
  \begin{minipage}{0.95\linewidth}
    \textbf{Prompts:} The ratio of the radii of two concentric circles is $1: 3$. The line $AC$ a diameter of the biggest circle. A chord $BC$ of the big circle touches the small circle (see diagram). The line $AB$ has length $12$. How big is the radius of the big circle?

    \medskip
    \textbf{Model Responses:} 

    \texttt{<think>}...\texttt{</think>} To solve the problem, we start by analyzing the given information and using geometric properties:  
    
    \subsection*{Step 1: Define Variables and Key Properties}  
    Let the center of the two concentric circles be \( O \). Let the radius of the \textbf{small circle} be \( r \), so the radius of the \textbf{big circle} is \( 3r \) (since their radii are in the ratio \( 1:3 \)).  
    
    \begin{itemize}
        \item \( AC \) is a diameter of the big circle, so \( AC = 2 \times 3r = 6r \).  
        \item \( BC \) is a chord of the big circle that is tangent to the small circle. For a tangent to a circle, the radius to the point of tangency is perpendicular to the tangent. Thus, if \( T \) is the point where \( BC \) touches the small circle, \( OT \perp BC \) and \( OT = r \) (radius of the small circle).  
    \end{itemize}
    
    \subsection*{Step 2: Use the Right Angle in Semicircle}  
    Since \( AC \) is a diameter of the big circle, by the \textbf{Thales’ theorem}, \( \angle ABC = 90^\circ \) (angle in a semicircle is a right angle). Thus, \( \triangle ABC \) is a right triangle with \( \angle ABC = 90^\circ \).  
    
    \subsection*{Step 3: Analyze \( \triangle OTB \)}  
    In \( \triangle OTB \):  
    \begin{itemize}
        \item \( OB = 3r \) (radius of the big circle).  
        \item \( OT = r \) (radius of the small circle).  
        \item \( OT \perp BC \), so \( \triangle OTB \) is a right triangle.  
    \end{itemize}
    
    By the Pythagorean theorem:  
    \[
    BT = \sqrt{OB^2 - OT^2} = \sqrt{(3r)^2 - r^2} = \sqrt{9r^2 - r^2} = \sqrt{8r^2} = 2r\sqrt{2}
    \]  
    Since \( OT \perp BC \) and \( O \) is the center of the circle, \( OT \) bisects \( BC \) (a perpendicular from the center to a chord bisects the chord). Thus \( BC = 2 \times BT = 2 \times 2r\sqrt{2} = 4r\sqrt{2} \).  
    
    \subsection*{Step 4: Apply Pythagoras to \( \triangle ABC \)}  
    In right triangle \( \triangle ABC \):  
    \[
    AB^2 + BC^2 = AC^2
    \]  
    We know \( AB = 12 \), \( BC = 4r\sqrt{2} \), and \( AC = 6r \). Substitute these values:  
    \[
    12^2 + (4r\sqrt{2})^2 = (6r)^2
    \]  
    Simplify:  
    \[
    144 + 32r^2 = 36r^2
    \]  
    Solve for \( r^2 \):  
    \[
    144 = 4r^2 \quad\Rightarrow\quad r^2 = 36 \quad\Rightarrow\quad r = 6 \ (\text{since radius is positive})
    \]  
    
    \subsection*{Step 5: Find the Radius of the Big Circle}  
    The radius of the big circle is \( 3r \). Substituting \( r = 6 \):  
    \[
    3r = 3 \times 6 = 18
    \]  
    
    Thus, the radius of the big circle is \(\mathbf{18}\).

  \end{minipage}

  \vspace{6pt}
  \noindent\color{black}\rule{\linewidth}{0.5pt}
  \captionsetup{font=small}
  \caption{An example illustrating the model's ability to solve geometric reasoning problems.}
\end{figure}

\subsection{Scientific Question Answering: Physics}

\begin{figure}[H]
  \centering
  \noindent\color{black}\rule{\linewidth}{0.5pt}
  \vspace{3pt}

  \includegraphics[width=0.55\linewidth,height=0.45\textheight,keepaspectratio]{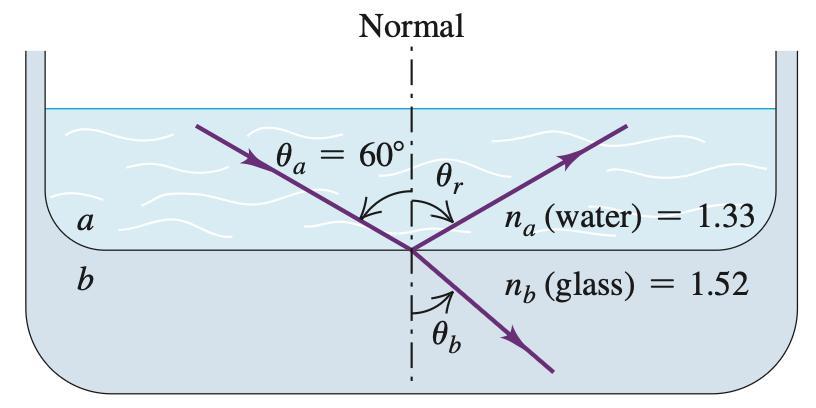}

  \vspace{6pt}
  \begin{minipage}{0.95\linewidth}
    \textbf{Prompts:} 	In figure, material \( a \) is water and material \( b \) is glass with index of refraction 1.52. find the directions of the reflected and refracted rays.

    \medskip
    \textbf{Model Responses:} 

    \texttt{<think>}...\texttt{</think>} To determine the directions of the reflected and refracted rays, we use the \textbf{law of reflection} for the reflected ray and \textbf{Snell’s Law} for the refracted ray.

    \subsubsection*{Step 1: Reflected Ray (Law of Reflection)}
    The law of reflection states that the angle of incidence (\(\theta_a\)) equals the angle of reflection (\(\theta_r\)):
    \[
    \theta_r = \theta_a
    \]
    Given \(\theta_a = 60^\circ\), the reflected ray makes an angle of \(60^\circ\) with the normal, on the same side as the incident ray.
    
    \subsubsection*{Step 2: Refracted Ray (Snell’s Law)}
    Snell’s Law relates the angles of incidence and refraction to the indices of refraction of the two media:
    \[
    n_a \sin\theta_a = n_b \sin\theta_b
    \]
    where:
    \begin{itemize}
      \item \(n_a = 1.33\) (index of refraction of water, material \(a\)),
      \item \(n_b = 1.52\) (index of refraction of glass, material \(b\)),
      \item \(\theta_a = 60^\circ\) (angle of incidence in water),
      \item \(\theta_b\) = angle of refraction in glass (what we need to find).
    \end{itemize}
    
    Rearrange Snell’s Law to solve for \(\sin\theta_b\):
    \[
    \sin\theta_b = \frac{n_a}{n_b} \sin\theta_a
    \]
    
    Substitute the known values:
    \[
    \sin\theta_b = \frac{1.33}{1.52} \cdot \sin(60^\circ)
    \]
    
    Calculate \(\sin(60^\circ) = \frac{\sqrt{3}}{2} \approx 0.8660\), then:
    \[
    \sin\theta_b = \frac{1.33}{1.52} \cdot 0.8660 \approx 0.875 \cdot 0.8660 \approx 0.758
    \]
    
    Take the inverse sine to find \(\theta_b\):
    \[
    \theta_b = \arcsin(0.758) \approx 49.3^\circ
    \]
    
    \subsubsection*{Final Directions}
    \begin{itemize}
      \item \textbf{Reflected ray}: Makes an angle of \(60^\circ\) with the normal, on the same side as the incident ray.
      \item \textbf{Refracted ray}: Makes an angle of approximately \(49.3^\circ\) with the normal, on the glass side (since \(n_b > n_a\), the ray bends \emph{toward} the normal).
    \end{itemize}
    
    Thus, the reflected ray has an angle of \(60^\circ\) with the normal, and the refracted ray has an angle of approximately \(49.3^\circ\) with the normal.

  \end{minipage}

  \vspace{6pt}
  \noindent\color{black}\rule{\linewidth}{0.5pt}
  \captionsetup{font=small}
  \caption{An example illustrating the model's ability to solve a physics refraction and reflection problem.}
\end{figure}

\subsection{Scientific Question Answering: Biology}

\begin{figure}[H]
  \centering
  \noindent\color{black}\rule{\linewidth}{0.5pt}
  \vspace{3pt}

  \includegraphics[width=0.65\linewidth,height=0.65\textheight,keepaspectratio]{figures/cases/Case_sci_bio2.png}

  \vspace{6pt}
  \begin{minipage}{0.95\linewidth}
    \textbf{Prompts:} The pedigree in shows the mode of inheritance of a human disease that is associated with mutations in Gene A. Note: Individuals marrying into this family DO NOT have the disease-associated allele unless shaded and the pedigree is completely penetrant. What is the genotype(s) of Individual 9?

    \medskip
    \textbf{Model Responses:} 

    \texttt{<think>}...\texttt{</think>} To determine the genotype of Individual 9, we first identify the \textbf{mode of inheritance} using the pedigree:  

    \subsubsection*{Step 1: Determine Mode of Inheritance}  
    \begin{itemize}
        \item Affected individuals are primarily males (shaded squares: 3, 7, 11, 18) and one female (shaded circle: 12).  
        \item In \textbf{X - linked recessive inheritance}, males are more frequently affected (they have only one X chromosome, so one recessive allele causes the disease), while females require two recessive alleles to be affected. This matches the pedigree:  
        \begin{itemize}
            \item Affected males inherit the recessive allele on their X chromosome from their mother (who may be a carrier).  
            \item Affected females (e.g., 12) inherit recessive alleles from both parents (father and mother).  
        \end{itemize}
    \end{itemize}
    
    \subsubsection*{Step 2: Analyze Genotypes of Parents of Individual 9}  
    Individual 9 is a female (unshaded, not affected) and a daughter of:  
    \begin{itemize}
        \item Individual 3 (affected male): Genotype = \( X^aY \) (has the recessive allele \( X^a \) on his single X chromosome).  
        \item Individual 4 (unshaded female): To produce an affected son (Individual 11, \( X^aY \)), Individual 4 must be a \textbf{carrier} (\( X^AX^a \))—she can pass \( X^a \) to her son.  
    \end{itemize}
    
    \subsubsection*{Step 3: Determine Genotype of Individual 9}  
    Individual 9 is a daughter of \( X^aY \) (Individual 3) and \( X^AX^a \) (Individual 4). When these parents reproduce:  
    \begin{itemize}
        \item Daughters inherit an X chromosome from each parent.  
        \item From Individual 3 (\( X^aY \)): Daughter inherits \( X^a \).  
        \item From Individual 4 (\( X^AX^a \)): Daughter can inherit \( X^A \) (dominant allele) or \( X^a \) (recessive allele).  
    \end{itemize}
    
    Since Individual 9 is \textbf{unshaded} (not affected), she cannot have two recessive alleles (\( X^aX^a \)). Thus, she must inherit \( X^A \) from Individual 4 and \( X^a \) from Individual 3.  
    
    Thus, the genotype of Individual 9 is \( \boldsymbol{X^AX^a} \).

  \end{minipage}

  \vspace{6pt}
  \noindent\color{black}\rule{\linewidth}{0.5pt}
  \captionsetup{font=small}
  \caption{An example illustrating the model's ability to determine genetic inheritance patterns in a biology problem.}
\end{figure}

\subsection{OCR}
\begin{figure}[H]
  \centering
  \noindent\color{black}\rule{\linewidth}{0.5pt}
  \vspace{3pt}

  \begin{minipage}[t]{0.46\textwidth}
    \vspace{0pt}
    \includegraphics[width=\linewidth,height=0.55\textheight,keepaspectratio]{figures/cases/case_ocr_chart_2.jpeg}
  \end{minipage}\hfill
  \begin{minipage}[t]{0.50\textwidth}
  \textbf{Prompts:} Find out the value of 'Calories/Energy per 100g/ml', 'Calories/Energy per serving', 'Calories/Energy \% daily value/intake', 'Total Fat per 100g/ml', 'Total Fat per serving', 'Total Fat \% daily value/intake', 'Carbohydrate per 100g/ml', 'Carbohydrate per serving', 'Carbohydrate \% daily value/intake', 'Protein per 100g/ml', 'Protein per serving', 'Protein \% daily value/intake', 'Serving Size' stated in the image. Extract the required information using the text in the image directly, and return the result in a dict with keys 'CE-P1', 'CE-PS', 'CE-D', 'TF-P1', 'TF-PS', 'TF-D', 'CAR-P1', 'CAR-PS', 'CAR-D', 'PRO-P1', 'PRO-PS', 'PRO-D', 'SS'. Your answer should be in the JSON format.
  
  \medskip
  \textbf{Model Responses:} 
\begin{verbatim}
{
    "CE-P1": "273kJ/65kcal",
    "CE-PS": "546kJ/130kcal",
    "CE-D": "7%",
    "TF-P1": "3.7g",
    "TF-PS": "7.4g",
    "TF-D": "11%",
    "CAR-P1": "4.7g",
    "CAR-PS": "9.5g",
    "CAR-D": "4%",
    "PRO-P1": "3.5g",
    "PRO-PS": "7.0g",
    "PRO-D": "14%",
    "SS": "200ml"
}
\end{verbatim}
  \end{minipage}

  \vspace{6pt}
  \noindent\color{black}\rule{\linewidth}{0.5pt}
  \captionsetup{font=small}
  \caption{An example illustrating the model's OCR capability.}
\end{figure}

\begin{figure}[H]
  \centering
  \noindent\color{black}\rule{\linewidth}{0.5pt}
  \vspace{3pt}

  \includegraphics[width=0.6\linewidth]{figures/cases/case_chart_ocr_4.png}

  \vspace{6pt}
  \begin{minipage}{0.95\linewidth}
    \textbf{Prompts:}  According to the information in the image, please pair the corresponding keys and values below: Keys that need to be paired are 'LOT \#', 'STUDY DIRECTOR /DEPARTMENT', 'NO. OF CONTAINERS', 'SPECIAL HANDLING REQUIREMENTS', 'PHYSICAL DESCRIPTION', 'GROSS WEIGHT', 'SHIPPED VIA', 'DATES'. Values that need to be paired are 'Ray David', '03 /01 /90', 'Clear brown liquid', 'Store refrigerated in dark.', 'Direct Exress', 'Not provided', '012590', 'One'. Extract the required information in the text, and return the result in a dict directly. Your answer should be in the JSON format.

    \medskip
    \textbf{Model Responses:} 
\begin{verbatim}
{
    "LOT #": "012590",
    "STUDY DIRECTOR /DEPARTMENT": "Ray David",
    "NO. OF CONTAINERS": "one",
    "SPECIAL HANDLING REQUIREMENTS": "Store refrigerated in dark.",
    "PHYSICAL DESCRIPTION": "Clear brown liquid",
    "GROSS WEIGHT": "Not provided",
    "SHIPPED VIA": "Direct Exress",
    "DATES": "03/01/90"
}
\end{verbatim} 
  \end{minipage}

  \vspace{6pt}
  \noindent\color{black}\rule{\linewidth}{0.5pt}
  \captionsetup{font=small}
  \caption{An example illustrating the model's OCR capability.}
\end{figure}

\begin{figure}[H]
  \centering
  \noindent\color{black}\rule{\linewidth}{0.5pt}
  \vspace{3pt}

  \includegraphics[width=0.6\linewidth]{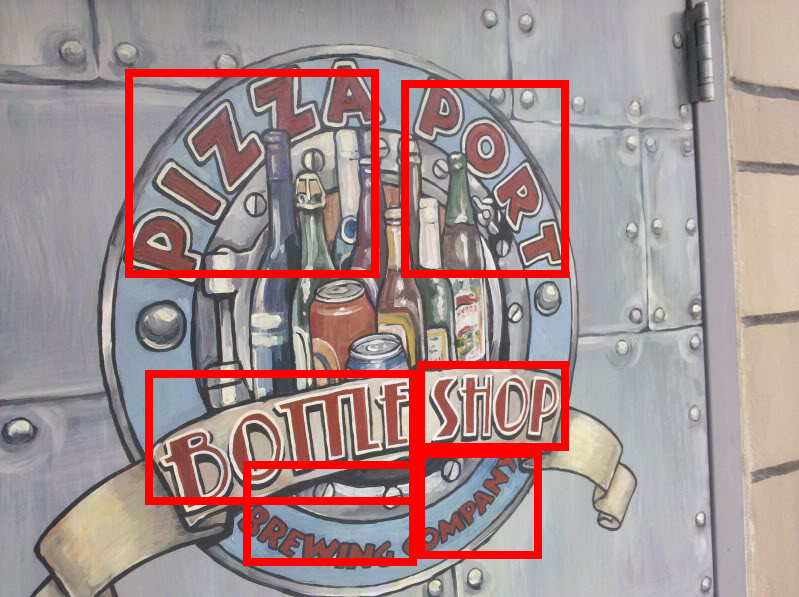}

  \vspace{6pt}
  \begin{minipage}{0.95\linewidth}
  \begin{flushleft}
    \textbf{Prompts:} Spot all the text in the image with word-level. Please provide the bounding box coordinates.

    \medskip
    \textbf{Model Responses:} 

    In the image, <ref>PIZZA</ref> is located at <box>(0.156,0.116),(0.474,0.466)</box>, <ref>PORT</ref> is located at <box>(0.502,0.134),(0.712,0.466)</box>, <ref>BOTTLE</ref> is located at <box>(0.182,0.620),(0.522,0.846)</box>, <ref>SHOP</ref> is located at <box>(0.522,0.604),(0.712,0.754)</box>, <ref>BREWING</ref> is located at <box>(0.304,0.772),(0.522,0.948)</box>, <ref>COMPANY</ref> is located at <box>(0.522,0.754),(0.678,0.936)</box> 
\end{flushleft}
  \end{minipage}

  \vspace{6pt}
  \noindent\color{black}\rule{\linewidth}{0.5pt}
  \captionsetup{font=small}
  \caption{An example illustrating the model's text localization capability. The boxes are not part of the original image; they were added to visualize the coordinates generated by the model.}
\end{figure}

\subsection{Chart}
\begin{figure}[H]
  \centering
  \noindent\color{black}\rule{\linewidth}{0.5pt}
  \vspace{3pt}

  \includegraphics[width=0.9\linewidth]{figures/cases/case_chart_1.png}

  \vspace{6pt}
  \begin{minipage}{0.95\linewidth}
    \textbf{Prompts:} Which country in Africa has a mean male height between 170 and 175cm but a child mortality rate between 10 and 15\%?

    \medskip
     \textbf{Model Responses:} \texttt{<think>}...\texttt{</think>}
To determine the African country meeting the criteria, follow these steps:

\begin{enumerate}
  \item \textbf{Identify the relevant axes:}
    \begin{itemize}
      \item \textit{X-axis}: Mean male height (170--175 cm).
      \item \textit{Y-axis}: Child mortality rate (10--15\%).
      \item \textit{Color}: Purple (represents ``Africa'' in the legend).
    \end{itemize}

  \item \textbf{Locate purple dots within the specified ranges:} \\
        Check for purple points where the x-value (height) is between 170 and 175 cm, and the y-value (child mortality) is between 10\% and 15\%.

  \item \textbf{Analyze the data point:} \\
        The country \textit{Senegal} has a mean male height of approximately 174 cm (within 170--175 cm) and a child mortality rate of approximately 14\% (within 10--15\%).
\end{enumerate}

Thus, the African country that meets the criteria is \textbf{Senegal}.
  \end{minipage}

  \vspace{6pt}
  \noindent\color{black}\rule{\linewidth}{0.5pt}
  \captionsetup{font=small}
  \caption{An example illustrating the model's ability to perform a conditional search on a chart.}
\end{figure}

\begin{figure}[H]
  \centering
  \noindent\color{black}\rule{\linewidth}{0.5pt}
  \vspace{3pt}

  \includegraphics[width=0.9\linewidth]{figures/cases/case_chart_2.png}

  \vspace{6pt}
  \begin{minipage}{0.95\linewidth}
    \textbf{Prompts:} Which Christmas cookie requires the most sugar?

    \medskip
     \textbf{Model Responses:} \texttt{<think>}...\texttt{</think>}
To determine which Christmas cookie requires the most sugar, we analyze the \textbf{flow lines} connecting ``Cookies'' to ``Sugar'' in the infographic. The thickness or prominence of the flow line indicates the amount of sugar used for each cookie.

\medskip

Among the listed cookies (Chocolate Crinkles, Snickerdoodles, PB Blossoms, Gingerbread, Peppermint Chip, Spritz), the flow line from \emph{Chocolate Crinkles} to ``Sugar'' is the most substantial. Additionally, Chocolate Crinkles are described as having a ``snowy powdered sugar coating,'' which implies a significant use of sugar.

\medskip

Thus, \textbf{Chocolate Crinkles} require the most sugar.
  \end{minipage}

  \vspace{6pt}
  \noindent\color{black}\rule{\linewidth}{0.5pt}
  \captionsetup{font=small}
  \caption{An example illustrating the model's ability to perform a comparative analysis on a flow chart.}
\end{figure}

\begin{figure}[H]
  \centering
  \noindent\color{black}\rule{\linewidth}{0.5pt}
  \vspace{3pt}

  \includegraphics[width=0.9\linewidth]{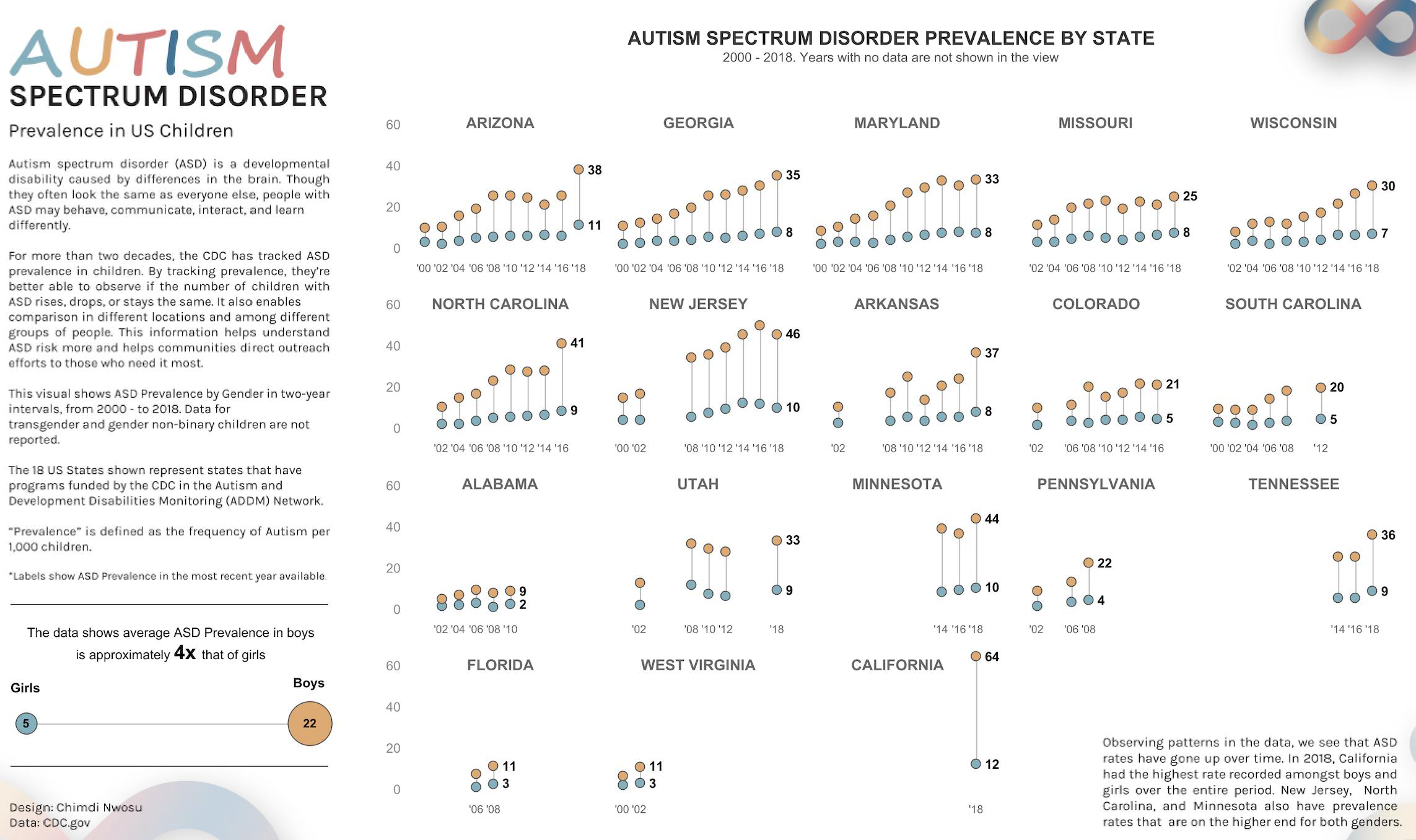}

  \vspace{6pt}
  \begin{minipage}{0.95\linewidth}
    \textbf{Prompts:} Looking at the CDC autism data from 2000-2018, list all states that have ever reported male autism prevalence rates (over the entire period) that were more than double South Carolina's male autism prevalence rate in 2012.

    \medskip
     \textbf{Model Responses:} \texttt{<think>}...\texttt{</think>}
To solve this, follow these steps:

\textbf{Step 1: Find South Carolina's male autism prevalence in 2012} \\
From the ``South Carolina'' graph, the \textbf{male} (\textit{orange dot}) prevalence in 2012 is 20.

\medskip
\textbf{Step 2: Calculate ``double'' South Carolina's 2012 male rate} \\
Double of 20 is \(20 \times 2 = 40\). We need states with \emph{any} male autism prevalence rate \textbf{greater than 40} over the 2000--2018 period.

\medskip
\textbf{Step 3: Analyze each state's male prevalence (orange dots)} \\
Examine the male prevalence (orange dots) for all states:
\begin{itemize}
  \item \textbf{North Carolina}: Male prevalence reaches 41 (exceeds 40).
  \item \textbf{New Jersey}: Male prevalence reaches 46 (exceeds 40).
  \item \textbf{Minnesota}: Male prevalence reaches 44 (exceeds 40).
  \item \textbf{California}: Male prevalence reaches 64 (exceeds 40).
\end{itemize}

Other states' maximum male prevalence (e.g., Arizona: 38; Georgia: 35; etc.) do not exceed 40.

\medskip
Thus, the states with male autism prevalence rates more than double South Carolina's 2012 rate are \textbf{North Carolina, New Jersey, Minnesota, and California}.

  \end{minipage}

  \vspace{6pt}
  \noindent\color{black}\rule{\linewidth}{0.5pt}
  \captionsetup{font=small}
  \caption{An example illustrating the model's ability to execute a multi-step analysis across a panel of charts.}
\end{figure}

\subsection{Grounding}

\begin{figure}[H]
  \centering
  \noindent\color{black}\rule{\linewidth}{0.5pt}
  \vspace{3pt}

  \includegraphics[width=0.6\linewidth]{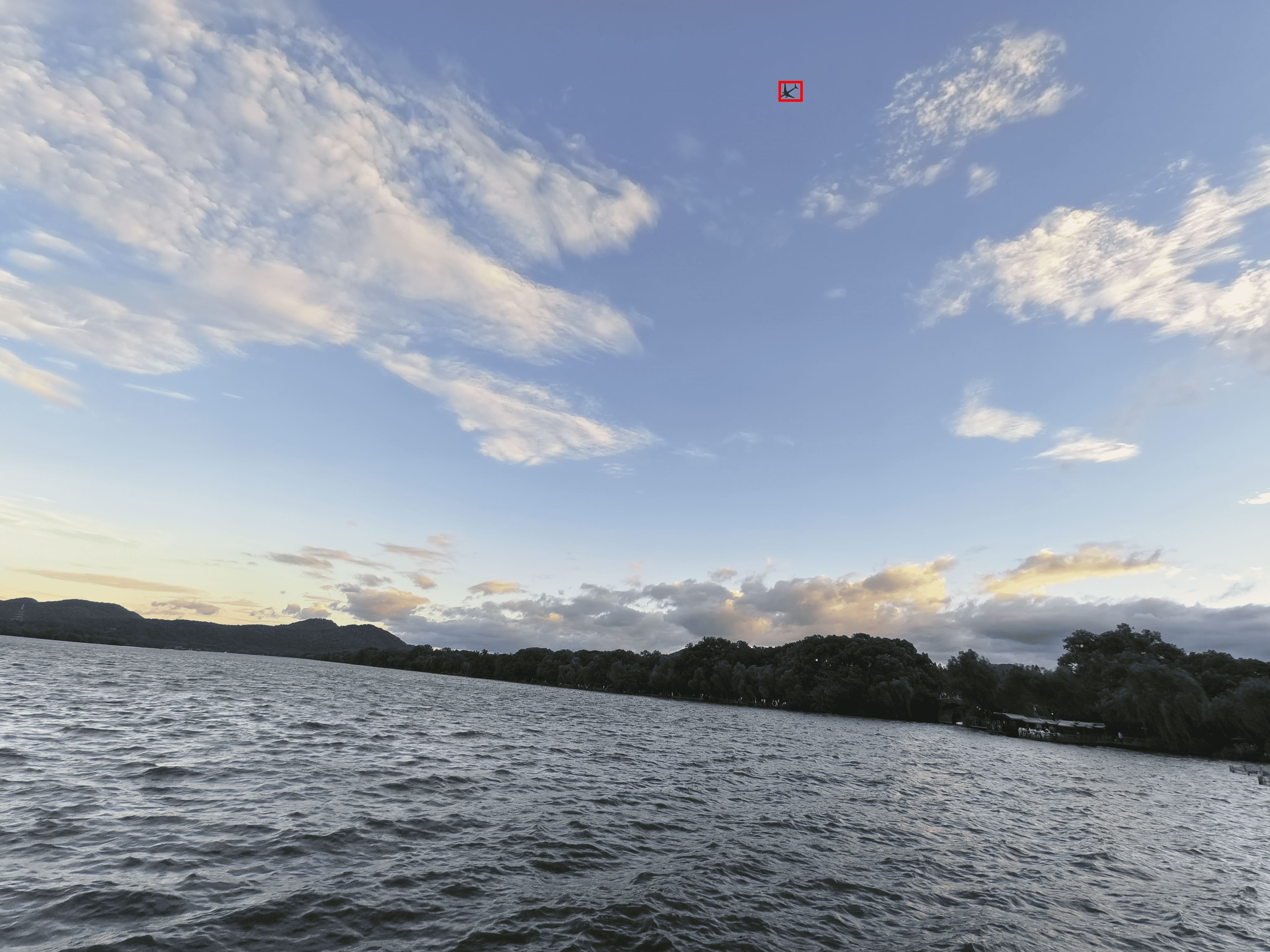}

  \vspace{6pt}
  \begin{minipage}{0.95\linewidth}
    \textbf{Prompts:} How can I locate the <ref>airplane</ref> in this image? Please provide the bounding box coordinates.

    \medskip
    \textbf{Model Responses:} 

    The <ref>airplane</ref> is found within \url{<box>(0.613,0.085),(0.632,0.107)</box>} of the image.
  \end{minipage}

  \vspace{6pt}
  \noindent\color{black}\rule{\linewidth}{0.5pt}
  \captionsetup{font=small}
  \caption{An example illustrating the model's ability to ground small object. The boxes are not part of the original image; they were added to visualize the coordinates generated by the model.}
\end{figure}

\begin{figure}[H]
  \centering
  \noindent\color{black}\rule{\linewidth}{0.5pt}
  \vspace{3pt}

  \includegraphics[width=0.6\linewidth]{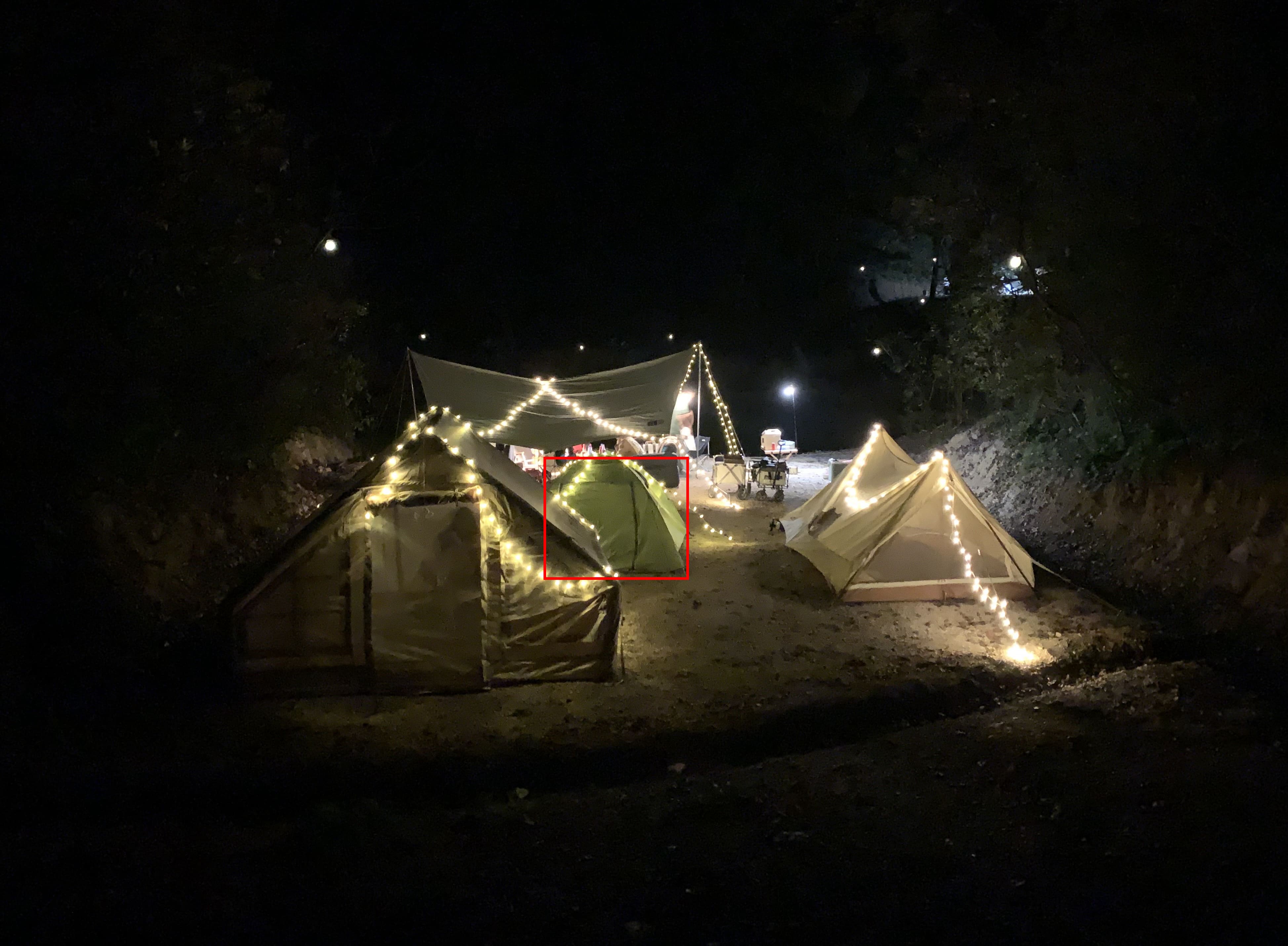}

  \vspace{6pt}
  \begin{minipage}{0.95\linewidth}
    \textbf{Prompts:} How can I locate the <ref>the green tent in the middle</ref> in this image? Please provide the bounding box coordinates.

    \medskip
    \textbf{Model Responses:} 
    The <ref>the green tent in the middle</ref> is positioned at \url{<box>(0.422,0.483),(0.535,0.613)</box>}.
  \end{minipage}

  \vspace{6pt}
  \noindent\color{black}\rule{\linewidth}{0.5pt}
  \captionsetup{font=small}
  \caption{An example illustrating the model's ability to ground referring object. The boxes are not part of the original image; they were added to visualize the coordinates generated by the model.}
\end{figure}

\begin{figure}[H]
  \centering
  \noindent\color{black}\rule{\linewidth}{0.5pt}
  \vspace{3pt}

  \includegraphics[width=0.6\linewidth]{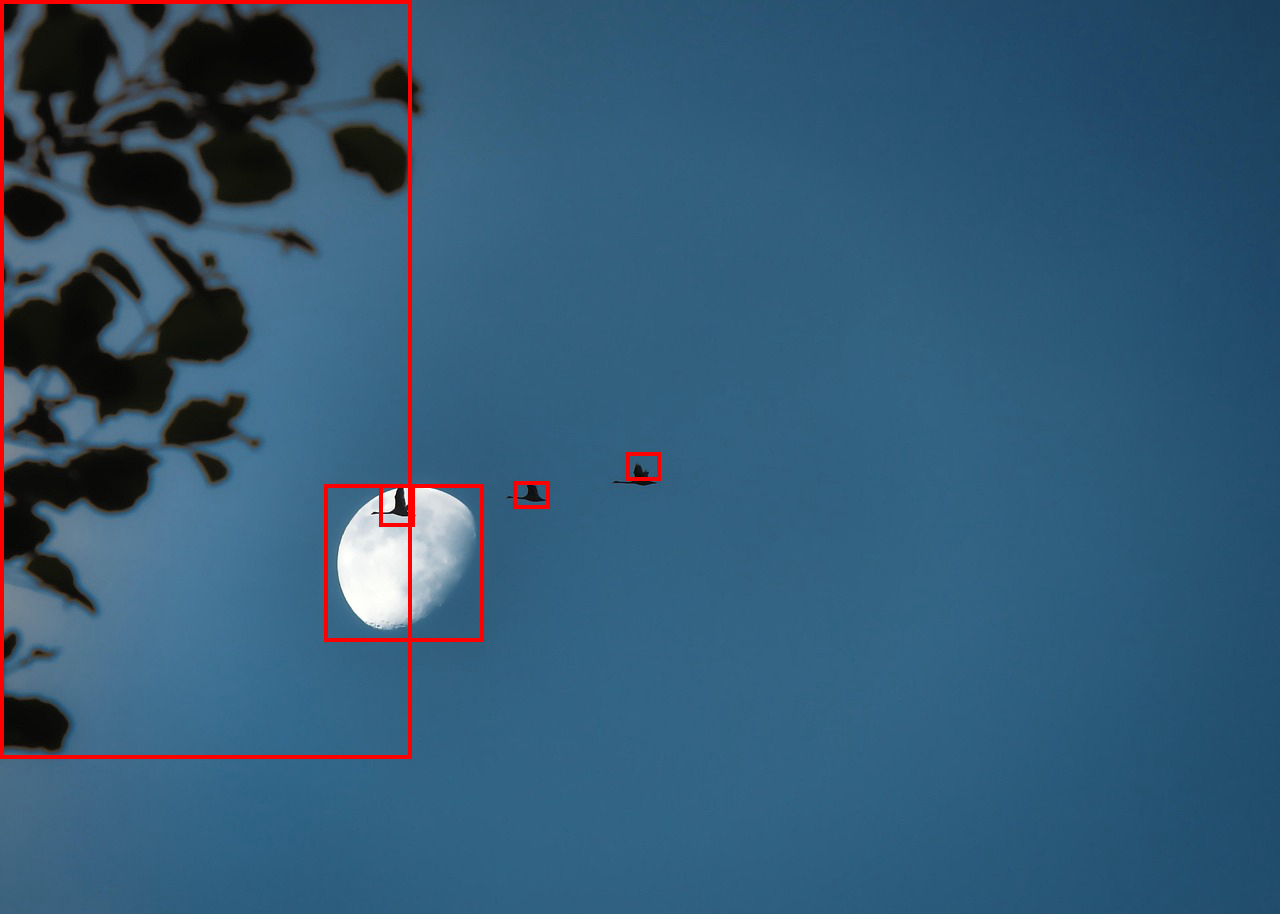}

  \vspace{6pt}
  \begin{minipage}{0.95\linewidth}
    \textbf{Prompts:} Describe this image with grounded objects. Please provide the bounding box coordinates.

    \medskip
    \textbf{Model Responses:} 
    The image features a serene scene with <ref>three birds</ref>\url{[<box>(0.401,0.526),(0.430,0.557)</box>,<box>(0.489,0.494),(0.516,0.526)</box>,<box>(0.296,0.529),(0.324,0.576)</box>]} flying in formation against a backdrop of a clear blue sky. The <ref>moon</ref>\url{<box>(0.253,0.529),(0.378,0.702)</box>} is prominently visible, partially obscured by one of the birds, creating a striking silhouette effect. In the upper left corner, there are <ref>tree branches with leaves</ref>\url{<box>(0.000,0.000),(0.322,0.830)</box>} that add depth and frame to the composition. The overall atmosphere is peaceful and evocative of a tranquil evening.
  \end{minipage}

  \vspace{6pt}
  \noindent\color{black}\rule{\linewidth}{0.5pt}
  \captionsetup{font=small}
  \caption{An example illustrating the model's ability to caption with grounding. The boxes are not part of the original image; they were added to visualize the coordinates generated by the model.}
\end{figure}

\begin{figure}[H]
  \centering
  \noindent\color{black}\rule{\linewidth}{0.5pt}
  \vspace{3pt}

  \includegraphics[width=0.6\linewidth]{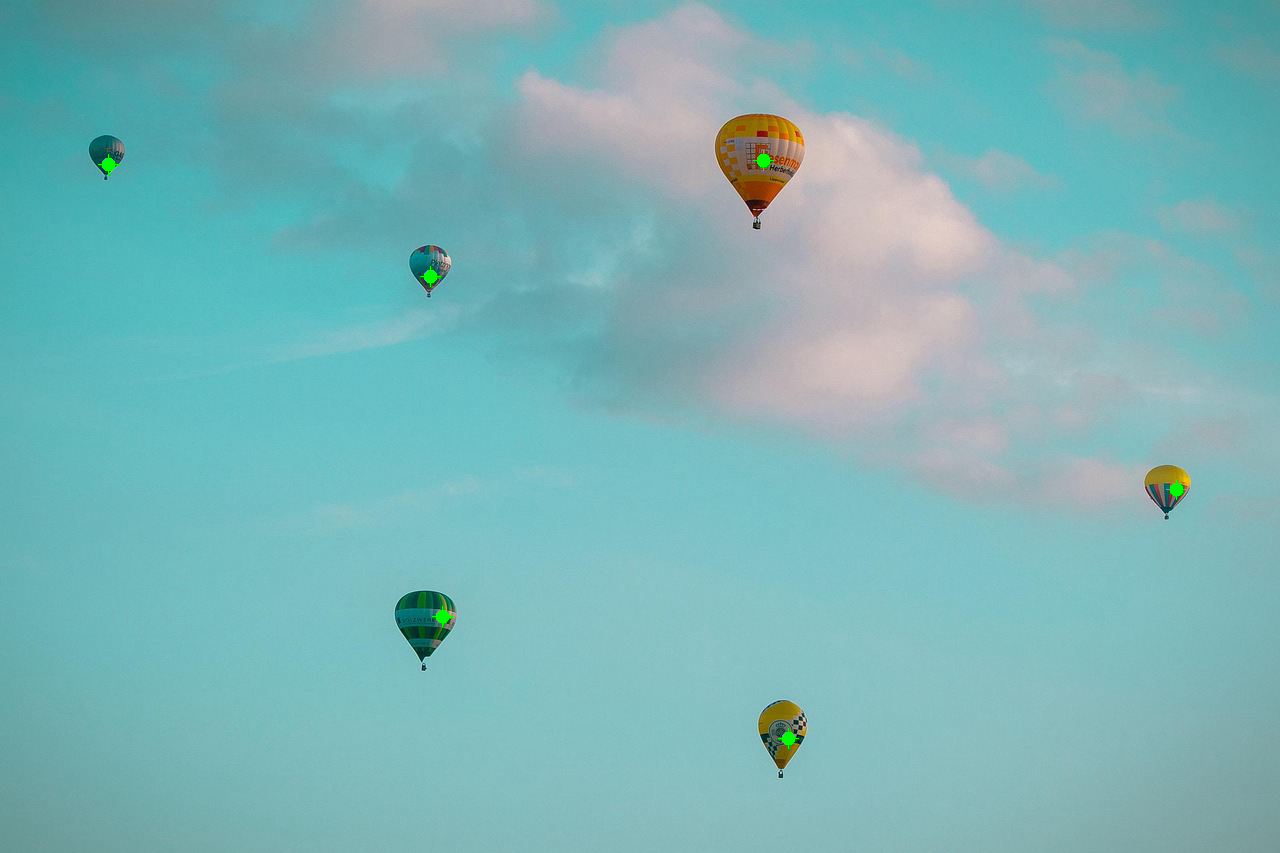}

  \vspace{6pt}
  \begin{minipage}{0.95\linewidth}
    \textbf{Prompts:} Where I can find <ref>hot air balloon</ref> in this image. Please provide the point coordinates.

    \medskip
    \textbf{Model Responses:} 

    The <ref>hot air balloon</ref> is located at \url{[<point>(0.084,0.192)</point>,<point>(0.336,0.323)</point>,<point>(0.345,0.722)</point>,<point>(0.596,0.188)</point>,<point>(0.616,0.865)</point>,<point>(0.919,0.573)</point>]}.
  \end{minipage}

  \vspace{6pt}
  \noindent\color{black}\rule{\linewidth}{0.5pt}
  \captionsetup{font=small}
  \caption{An example illustrating the model's ability to ground with point. The points are not part of the original image; they were added to visualize the coordinates generated by the model.}
\end{figure}

\subsection{Multi-Images}

\begin{figure}[H]
  \centering
  \noindent\color{black}\rule{\linewidth}{0.5pt}
  \vspace{3pt}
  \includegraphics[width=0.95\linewidth,height=0.45\textheight,keepaspectratio,trim=0 0 0 -20, clip]{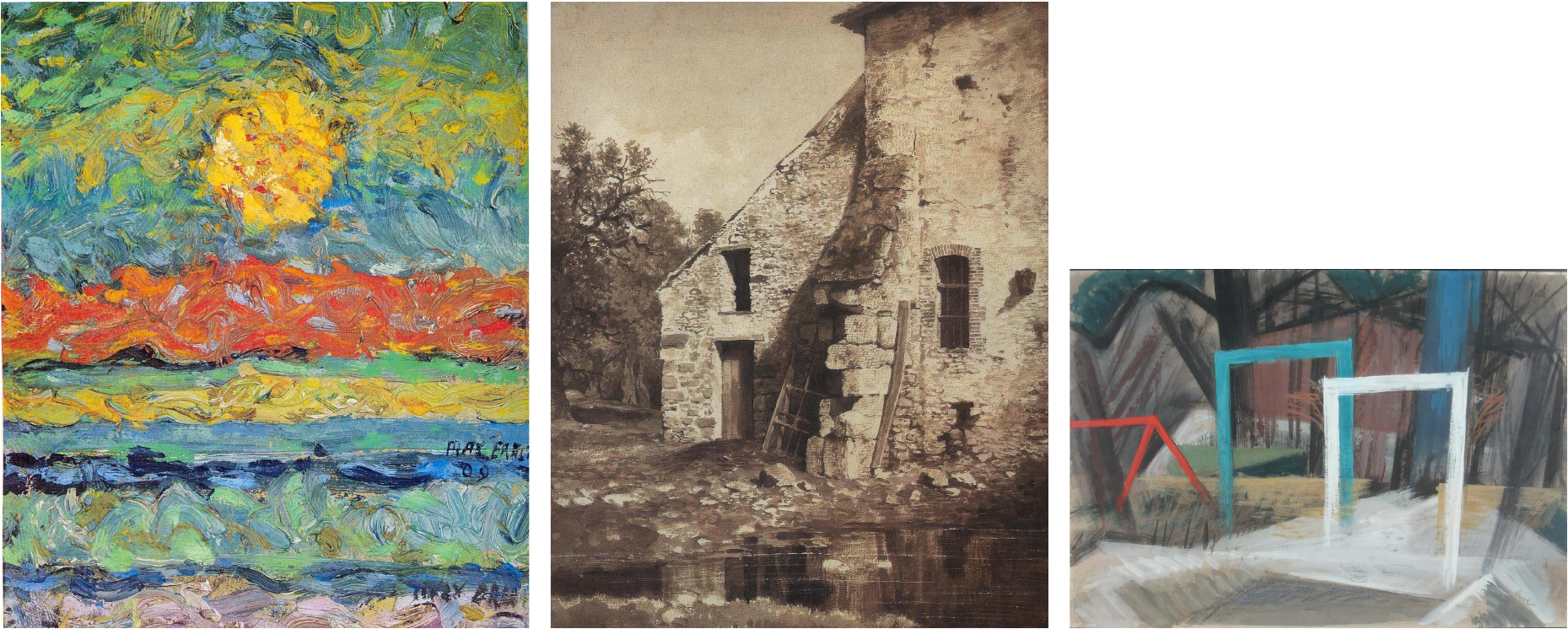}

  \vspace{6pt}
  \begin{minipage}{0.95\linewidth}
    \textbf{Prompts:} Given the following images of art paintings, use the first image as the reference image, and determine which one of the second or the third image shares the same style as the reference image?

    \medskip
    \textbf{Model Responses:} 

    \texttt{<think>}...\texttt{</think>} The first image, with its vibrant and dynamic brushstrokes, is characteristic of the Post-Impressionist style, particularly reminiscent of Vincent van Gogh's work. The second image, on the other hand, is a more subdued and realistic depiction of a rural scene, likely belonging to the Realist or Naturalist school of art. The third image, however, shares a similar style to the first image, with its bold use of color, expressive brushwork, and abstract elements. The third image's abstract forms and vibrant color palette align more closely with the Post-Impressionist style seen in the first image. Therefore, the third image shares the same style as the reference image.
  \end{minipage}

  \vspace{6pt}
  \noindent\color{black}\rule{\linewidth}{0.5pt}
  \captionsetup{font=small}
  \caption{An example illustrating the model's ability to recognize art style.}
\end{figure}

\begin{figure}[H]
  \centering
  \noindent\color{black}\rule{\linewidth}{0.5pt}
  \vspace{3pt}
  \includegraphics[width=0.95\linewidth,height=0.45\textheight,keepaspectratio,trim=0 0 0 -20, clip]{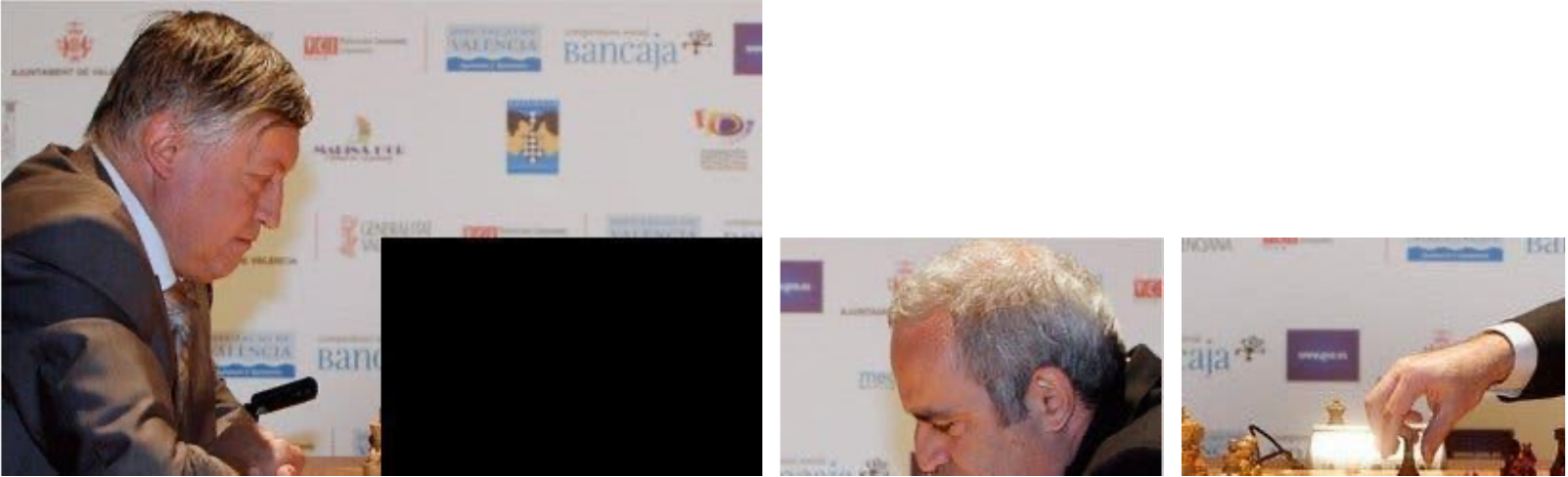}

  \vspace{6pt}
  \begin{minipage}{0.95\linewidth}
    \textbf{Prompts:} Given the first image with the lower right corner missing, can you tell which one of the second image or the third image is the missing part?

    \medskip
    \textbf{Model Responses:} 

    \texttt{<think>}...\texttt{</think>} The third image is the missing part of the first image.
  \end{minipage}

  \vspace{6pt}
  \noindent\color{black}\rule{\linewidth}{0.5pt}
  \captionsetup{font=small}
  \caption{An example illustrating the model's ability on the jigsaw challenge.}
\end{figure}

\subsection{Video}

\begin{figure}[H]
  \centering
  \noindent\color{black}\rule{\linewidth}{0.5pt}
  \vspace{3pt}
  \includegraphics[width=0.95\linewidth,height=0.45\textheight,keepaspectratio,trim=0 0 0 -20, clip]{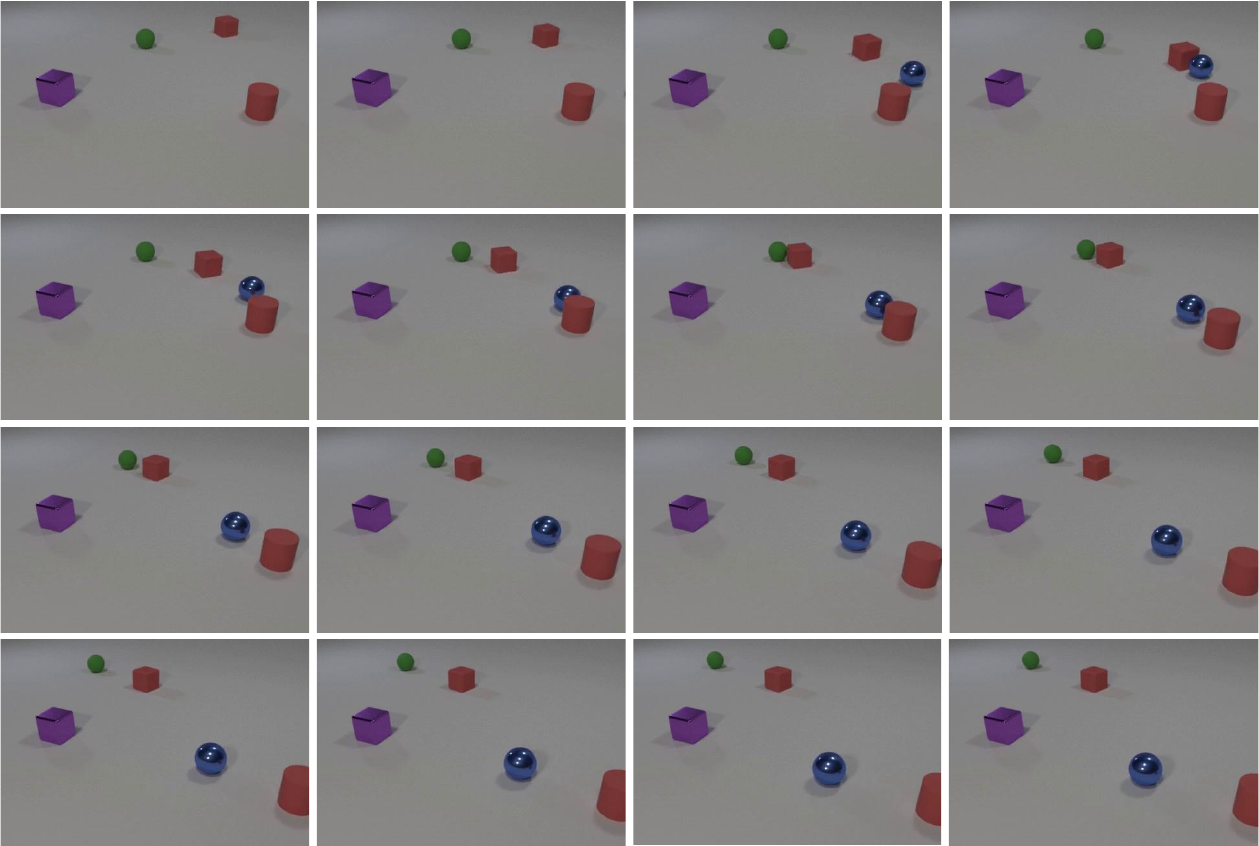}

  \vspace{6pt}
  \begin{minipage}{0.95\linewidth}
    \textbf{Prompts:} Describe the collision process of the blue sphere in the video.

    \medskip
    \textbf{Model Responses:} 

    \texttt{<think>}...\texttt{</think>} The blue sphere collides with the red cube, then collides with the red cylinder.
  \end{minipage}

  \vspace{6pt}
  \noindent\color{black}\rule{\linewidth}{0.5pt}
  \captionsetup{font=small}
  \caption{An example illustrating the model's ability on video perception and reasoning.}
\end{figure}

\begin{figure}[H]
  \centering
  \noindent\color{black}\rule{\linewidth}{0.5pt}
  \vspace{3pt}
  \includegraphics[width=0.95\linewidth,height=0.45\textheight,keepaspectratio,trim=0 0 0 -20, clip]{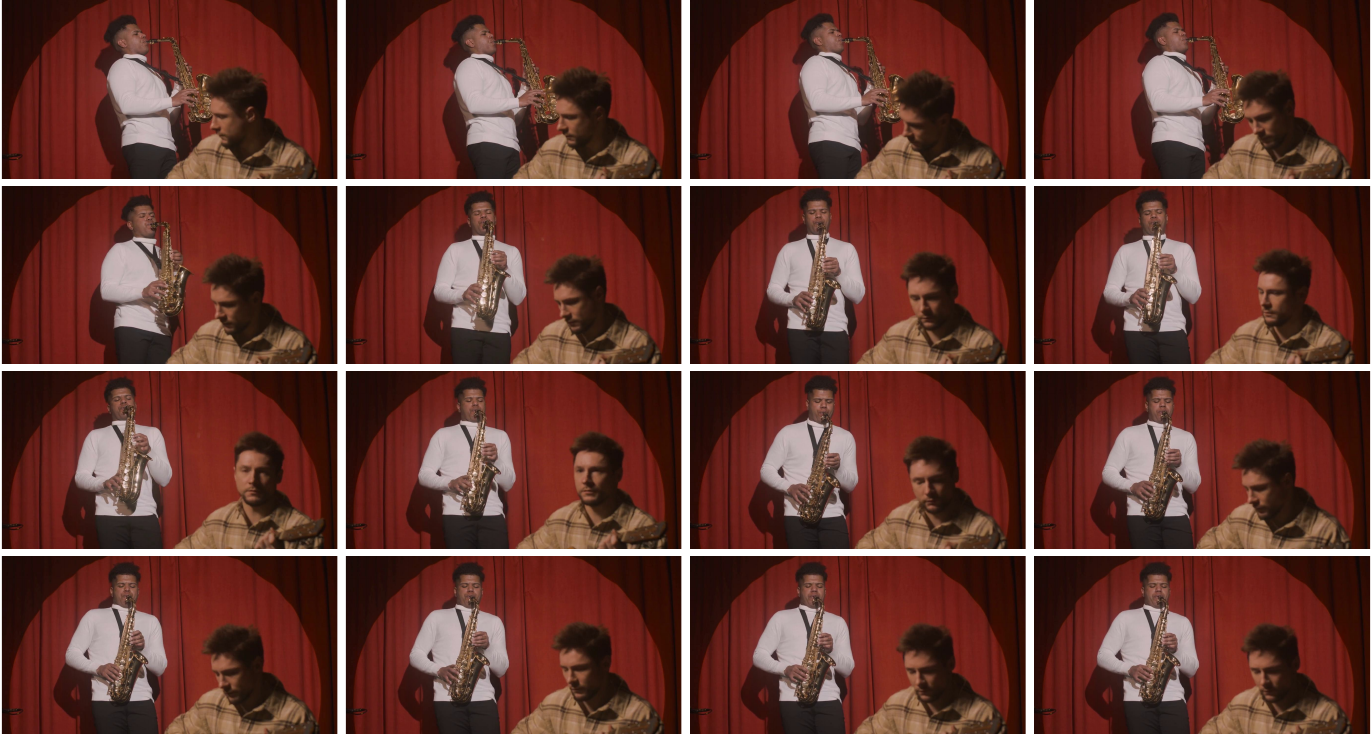}

  \vspace{6pt}
  \begin{minipage}{0.95\linewidth}
    \textbf{Prompts:} Describe the video content in detail.

    \medskip
    \textbf{Model Responses:} 

    \texttt{<think>}...\texttt{</think>} The video captures a live musical performance featuring two musicians on stage, set against a backdrop of a red curtain that creates a warm and intimate atmosphere. The primary focus is on a saxophonist dressed in a white long-sleeved shirt and black pants, who is actively engaged in playing the saxophone. His posture and the positioning of his fingers on the instrument suggest a dynamic and expressive performance. The saxophonist's shadow is cast on the red curtain behind him, adding depth to the scene.

    In the foreground, slightly out of focus, is a guitarist wearing a plaid shirt, who is also deeply involved in the performance. The guitarist's presence, along with the saxophonist's, contributes to the overall ambiance of a live musical event. The lighting is strategically focused on the performers, highlighting their actions and creating a spotlight effect that enhances the visual appeal of the scene.
    
    Throughout the video, the saxophonist's performance remains the central element, with his posture and the positioning of his fingers on the saxophone indicating a continuous and expressive musical piece. The guitarist's role is secondary but essential, as his presence complements the saxophonist's performance, adding layers to the musical experience.
    
    The video maintains a consistent visual theme, with the red curtain backdrop and focused lighting on the performers creating a cohesive and engaging atmosphere. The saxophonist's shadow on the curtain adds a visual element that enhances the depth and dimension of the scene. The overall composition of the video emphasizes the live musical performance, capturing the essence of the musicians' engagement and the intimate setting of the stage.
  \end{minipage}

  \vspace{6pt}
  \noindent\color{black}\rule{\linewidth}{0.5pt}
  \captionsetup{font=small}
  \caption{An example illustrating the model's ability to generate a detailed description for a given video.}
\end{figure}

\end{document}